\documentclass{article}

\usepackage{PRIMEarxiv}

\usepackage[square,numbers,compress]{natbib}
\usepackage[utf8]{inputenc} 
\usepackage[T1]{fontenc}    
\usepackage{hyperref}       
\usepackage{url}            
\usepackage{booktabs}       
\usepackage{amsfonts}       
\usepackage{nicefrac}       
\usepackage{microtype}      
\usepackage{xcolor}         
\usepackage{graphicx}
\usepackage{wrapfig}
\usepackage{amsmath}
\usepackage{amssymb}
\usepackage{mathtools}
\usepackage{amsthm}
\usepackage{bm}
\usepackage{pifont}
\usepackage{enumitem}
\usepackage{color, colortbl}
\usepackage{subcaption}
\usepackage{graphics}
\usepackage{caption}
\usepackage{wrapfig}
\usepackage[export]{adjustbox}
\usepackage{algorithm}
\usepackage{algorithmic}
\usepackage{multirow}

\def\eqref#1{equation~\ref{#1}}

\def\1{\bm{1}}

\def\vtheta{{\bm{\theta}}}

\def\vw{{\bm{w}}}
\def\vx{{\bm{x}}}

\DeclareMathAlphabet{\mathsfit}{\encodingdefault}{\sfdefault}{m}{sl}
\SetMathAlphabet{\mathsfit}{bold}{\encodingdefault}{\sfdefault}{bx}{n}

\def\sR{{\mathbb{R}}}

\newcommand{\E}{\mathbb{E}}

\graphicspath{ {./img/} }

\title{Score-based Conditional Generation with Fewer Labeled Data by Self-calibrating Classifier Guidance
}

\author{
  Paul Kuo-Ming Huang \\
  National Taiwan University \\
  \texttt{b08902072@csie.ntu.edu.tw} \\
  \And
  Si-An Chen \\
  National Taiwan University \\
  \texttt{d09922007@csie.ntu.edu.tw} \\
  \AND
  Hsuan-Tien Lin \\
  National Taiwan University \\
  \texttt{htlin@csie.ntu.edu.tw} \\
}

\begin{document}
\maketitle

\begin{abstract}
Score-based generative models (SGMs) are a popular family of deep generative models that achieve leading image generation quality. Early studies extend SGMs to tackle class-conditional generation by coupling an unconditional SGM with the guidance of a trained classifier. Nevertheless, such classifier-guided SGMs do not always achieve accurate conditional generation, especially when trained with fewer labeled data. We argue that the problem is rooted in the classifier's tendency to overfit without coordinating with the underlying unconditional distribution. To make the classifier respect the unconditional distribution, we propose improving classifier-guided SGMs by letting the classifier regularize itself. The key idea of our proposed method is to use principles from energy-based models to convert the classifier into another view of the unconditional SGM. Existing losses for unconditional SGMs can then be leveraged to achieve regularization by calibrating the classifier's internal unconditional scores. The regularization scheme can be applied to not only the labeled data but also unlabeled ones to further improve the classifier. Across various percentages of fewer labeled data, empirical results show that the proposed approach significantly enhances conditional generation quality. The enhancements confirm the potential of the proposed self-calibration technique for generative modeling with limited labeled data.
\end{abstract}

\section{Introduction}
Score-based generative models (SGMs) capture the underlying data distribution by learning the gradient function of the log-likelihood on data, also known as the score function. SGMs, when coupled with a diffusion process that gradually converts noise to data, can often synthesize higher-quality images than other popular alternatives, such as generative adversarial networks~\citep{song2021scorebased, dhariwal2021diffusion}.
The community's research dedication to SGMs demonstrates promising performance in image generation~\citep{song2021scorebased} and other fields such as audio synthesis~\citep{kong2021diffwave, jeong21_interspeech, huang2022fastdiff} and natural language generation~\citep{li2022diffusionlm}. 
Many such successful SGMs focus on unconditional generation, which models the distribution without considering other variables~\citep{song2019generative, ho2020denoising, song2021scorebased}. 
When seeking to generate images controllably from a particular class, it is necessary to model the conditional distribution concerning another variable. Such \textit{conditional} SGMs~\citep{song2021scorebased, dhariwal2021diffusion, chao2022denoising} will be the focus of this paper.

There are two major families of conditional SGMs. Classifier-free SGMs (CFSGMs) adopt specific conditional network architectures and losses~\citep{dhariwal2021diffusion, ho2021classifierfree}.
CFSGMs are known to generate high-fidelity images
when all data are labeled.
Nevertheless, our findings indicate that their performance drops significantly as the proportion of labeled data decreases. This disadvantage makes them less preferable in the semi-supervised setting, which is a realistic scenario when obtaining class labels takes significant time and costs.
Classifier-guided SGMs (CGSGMs) form another family of conditional SGMs~\citep{song2021scorebased, dhariwal2021diffusion} based on decomposing the conditional score into the unconditional score plus the gradient of an auxiliary classifier. A vanilla CGSGM can then be constructed by learning a classifier in parallel to training an unconditional SGM with the popular denoising score matching~\citep[DSM;][]{Vincent2011Denoise} technique.
The classifier 
can control the trade-off between generation diversity and fidelity~\citep{dhariwal2021diffusion}. 
Furthermore, because the unconditional SGM can be trained with both labeled and unlabeled data in principle, 
CGSGMs emerge with more potential than CFSGMs for the semi-supervised setting with fewer labeled data.

\begin{wrapfigure}{R}{0.4\linewidth}
\vspace{-0.4cm}
\centering
\begin{subfigure}{0.7\linewidth}
\includegraphics[width=\linewidth]{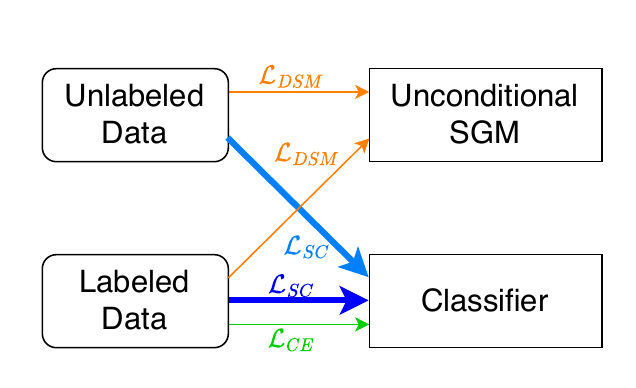}
\end{subfigure}
\caption{Illustration of proposed approach. A vanilla CGSGM takes the orange (DSM loss) and green (cross-entropy loss) arrows. The proposed CGSGM-SC additionally considers the two blue arrows representing the proposed self-calibration loss on both labeled and unlabeled data.}
\label{fig:cg_illus}
\vspace{-0.4cm}
\end{wrapfigure}

The quality of the classifier gradients is critical for CGSGMs. If the classifier 
overfits~\citep{lee2018training, NEURIPS2019_f1748d6b, NEURIPS2020_aeb7b30e, Grathwohl2020Your} and predicts highly inaccurate gradients, the resulting conditional scores may be unreliable, which lowers the generation quality even if the reliable unconditional scores can ensure decent generation fidelity.
Although there are general regularization techniques~\citep{zhang2018three, NEURIPS2019_f1748d6b, hoffman2019robust} that mitigate overfitting, their specific benefits for CGSGMs have not been fully studied except for a few cases~\citep{https://doi.org/10.48550/arxiv.2208.08664}. In fact, we find that those techniques are often not aligned with the unconditional SGM's view of the underlying distribution and offer limited benefits for improving CGSGMs.
One pioneering enhancement of CGSGM on distribution alignment, denoising likelihood score matching~\citep[CG-DLSM;][]{chao2022denoising}, calibrates the classifier with a regularization loss that aligns the classifier's gradients to the ground truth gradients with the \textit{external} help of unconditional SGMs.
Despite being able to achieve state-of-the-art performance within the CGSGM family, CG-DLSM is only designed for labeled data and does not apply to unlabeled data.

In this work, we design a regularization term that calibrates the classifier \textit{internally}, without relying on the unconditional SGM. Such an internal regularization has been previously achieved by the joint energy-based model~\citep[JEM;][]{Grathwohl2020Your}, which interprets classifiers as energy-based models. The interpretation allows JEM to define an auxiliary loss term that respects the underlying distribution and can unlock the generation capability of classifiers when using MCMC sampling. Nevertheless, extending JEM as CGSGM is non-trivial, as the sampling process is time-consuming and results in unstable loss values when coupled with the diffusion process of CGSGM. We thus take inspiration from JEM to derive a novel CGSGM regularizer instead of extending JEM directly.

Our design broadens the JEM interpretation of classifiers to be unconditional SGMs. Then, a stable and efficient self-calibration (SC) loss (as illustrated with $\mathcal{L}_{\mathrm{SC}}$ in Fig.~\ref{fig:cg_illus}) can be computed from the classifier \textit{internally} for regularization. The SC loss inherits a sound theoretical guarantee from the DSM technique for training unconditional SGMs. Our proposed CGSGM-SC approach, as shown in Fig.~\ref{fig:cg_illus}, allows separate training of the unconditional SGM and the classifier. The approach applies the SC loss on both labeled and \textit{unlabeled} data, resulting in immediate advantages in the semi-supervised setting with fewer labeled data.

Following earlier studies on CGSGMs~\citep{chao2022denoising}, we visually study the effectiveness of CGSGM-SC on a synthesized data set.
The results reveal that the CGSGM-SC leads to more accurate classifier gradients than vanilla CGSGM, thus enhancing the estimation of conditional scores. 
We further conduct thorough experiments on CIFAR-10 and CIFAR-100 datasets to validate the advantages of CGSGM-SC. The results confirm that CGSGM-SC is superior to the vanilla CGSGM and the state-of-the-art CGSGM-DLSM approach. Furthermore, in an extreme setting for which only $5\%$ of the data is labeled, CGSGM-SC, which more effectively utilizes unlabeled data, is significantly better than all CGSGMs and CFSGMs. This confirms the potential of CGSGM-SC in scenarios where labeled data are costly to obtain.

\section{Background}
\label{sec:back}
Consider a data distribution $p(\vx)$ where $\vx\in \sR^d$. The purpose of an SGM is to generate samples from $p(\vx)$ via the information contained in the score function $\nabla_\vx\log p(\vx)$, which is learned from data.
We first introduce how a diffusion process can be combined with learning a score function to effectively sample from $p(\vx)$ in Section~\ref{sec:back_sde}. Next, a comprehensive review of works that have extended SGMs to conditional SGMs is presented in Section~\ref{sec:classguide}, including those that incorporates classifier regularization for CGSGMs. Finally, JEM~\citep{Grathwohl2020Your} is introduced in Section~\ref{sec:jem}, highlighting its role in inspiring our proposed methodology.

\subsection{Score-based Generative Modeling by Diffusion}
\label{sec:back_sde}
\citet{song2021scorebased} propose to model the transition from a known prior distribution $p_T(\vx)$, typically a multivariate gaussian noise, to an unknown target distribution $p_0(\vx)=p(\vx)$ using the markov chain described by the following stochastic differential equation (SDE):
\begin{equation}
    d\vx=\bigl[f(\vx,t)-g(t)^2 s(\vx, t)\bigr]dt+g(t)d\bar{\vw},
    \label{eqn:revsde}
\end{equation}
where $\bar{\vw}$ is a standard Wiener process when the timestep flows from $T$ back to~$0$, $s(\vx, t) = \nabla_\vx\log p_t(\vx)$ denotes a time-dependent score function, and $f(\vx,t)$ and $g(t)$ are some prespecified functions that describe the overall movement of the distribution $p_t(\vx)$. The score function is learned by optimizing the following time-generalized denoise score matching (DSM)~\citep{Vincent2011Denoise} loss
\begin{align}
    \label{eqn:dsm}
    &\mathcal{L}_{DSM}(\vtheta)\nonumber\\&=\E_t\left[\lambda(t)\E_{\vx_t,\vx_0}\left[\frac{1}{2}\left\lVert s(\vx_t,t;\vtheta) - s_t(\vx_t|\vx_0)\right\rVert_2^2\right]\right],
\end{align}
where $t$ is selected uniformly between $0$ and~$T$, $\vx_t\sim p_t(\vx)$, $\vx_0\sim p_0(\vx)$, $s_t(\vx_t|\vx_0)$ denotes the score function of the noise distribution $p_t(\vx_t|\vx_0)$, which can be calculated using the prespecified $f(\vx,t)$ and $g(t)$, and $\lambda(t)$ is a weighting function that balances the loss of different timesteps.
In this paper, we use the hyperparameters from the original VE-SDE framework~\citep{song2021scorebased}.
A more detailed introduction to learning the score function and sampling through SDEs is described in Appendix~\ref{appendix:more_on_sde}.

\subsection{Conditional Score-based Generative Models}
\label{sec:classguide}
In conditional SGMs, we are given labeled data $\{(\vx_{m}, y_{m})\}_{m=1}^M$ in addition to unlabeled data $\{\vx_n\}_{n=M+1}^{M+N}$, where $y_m \in \{1, 2, \ldots, K\}$ denotes the class label. The case of $N = 0$ is called the fully-supervised setting; in this paper, we consider the semi-supervised setting with $N > 0$, with a particular focus on the challenging scenario where $\frac{M}{N+M}$ is small.
The goal of conditional SGMs is to learn the conditional score function $\nabla_\vx \log p(\vx | y)$ and then generate samples from $p(\vx | y)$, typically using a diffusion process as discussed in Section~\ref{sec:back_sde} and Appendix~\ref{appendix:generate_from_sde}.

One approach for conditional SGMs is classifier-free SGM~\citep{dhariwal2021diffusion, ho2021classifierfree}, which parameterizes its model with a joint architecture such that the class labels $y$ can be included as inputs. Classifier-free guidance~\citep{ho2021classifierfree}, also known as CFG, additionally uses a null token $y_{\textsc{nil}}$ to indicate unconditional score calculation, which is linearly combined with conditional score calculation for some specific $y$ to form the final estimate of $s(\vx | y)$. CFG is a state-of-the-art conditional SGM in the fully-supervised setting. Nevertheless, as we shall show in our experiments, its performance drops significantly in the semi-supervised setting, as the conditional parts of CFG may lack sufficient labeled data during training.

Another popular family of conditional SGM is CGSGM.
Under this framework, we decompose the conditional score function using Bayes' theorem~\citep{song2021scorebased, dhariwal2021diffusion}:
\begin{align}\label{eqn:class_guide}
    \nabla_\vx \log p(\vx|y)&=\nabla_\vx[\log p(\vx) + \log p(y|\vx)- \log p(y)]\nonumber \\
    &=\nabla_\vx\log p(\vx) + \nabla_\vx\log p(y|\vx)
\end{align}
The $\log p(y)$ term can be dropped because it is not a function of $\vx$ and is thus of gradient~$0$. The decomposition shows that conditional generation can be achieved by an unconditional SGM that learns the score function $\nabla_\vx \log p(\vx)$ plus an extra conditional gradient term $\nabla_\vx\log p(y|\vx)$.

The vanilla classifier-guidance (CG) estimates $\nabla_\vx\log p(y|\vx)$ with an auxiliary classifier trained from the cross-entropy loss on the labeled data and learns the unconditional score function by the denoising score matching loss $\mathcal{L}_{\mathrm{DSM}}$, which in principle can be applied on unlabeled data along with labeled data.
Nevertheless, the classifier within the vanilla CG approach is known to be potentially overconfident~\citep{lee2018training, NEURIPS2019_f1748d6b, NEURIPS2020_aeb7b30e, Grathwohl2020Your} in its predictions, which in turn results in inaccurate gradients. This can mislead the conditional generation process and decrease class-conditional generation quality.

\citet{dhariwal2021diffusion} propose to address the issue by post-processing the term $\nabla_\vx\log p(y|\vx)$ with a scaling parameter $\lambda_{CG} \neq 1$.
\begin{align}
\label{eqn:cg_scale}
    \nabla_\vx \log p(\vx|y)&= s(\vx) + \lambda_{CG}\nabla_\vx\log p(y|\vx; \pmb{\phi}),
\end{align}
where $s(\vx)=\nabla_\vx\log p(\vx)$ is the score function and $p(y|\vx; \pmb{\phi})$ is the posterior probability distribution outputted by a classifier parameterized by $\pmb{\phi}$. Increasing $\lambda_{CG}$ sharpens the distribution $p(y | \vx; \pmb{\phi})$, guiding the generation process to produce less diverse but higher fidelity samples. While the tuning heuristic is effective in improving the vanilla CG approach, it is not backed by sound theoretical explanations.

Other attempts to regularize the classifier \textit{during training} for resolving the issue form a promising research direction.
For instance, CGSGM with denoising likelihood score matching~\citep[CG-DLSM;][]{chao2022denoising} presents a regularization technique that employs the DLSM loss below formulated from the classifier gradient $\nabla_\vx\log p(y,t|\vx; \pmb{\phi})$ and unconditional score function $s_t(\vx)$. 
\vspace{-0.2cm}
\begin{align}
\label{eqn:dlsm}
    \mathcal{L}_{\mathrm{DLSM}}(\pmb{\phi})=
    &\E_t\left[\lambda(t)\E_{\vx_t,\vx_0}\left[\frac{1}{2}\left\lVert \nabla_\vx\log p(y,t|\vx_t; \pmb{\phi}) \right. \right. \right. \nonumber\\ &\left. \left. \vphantom{\frac{1}{2}}\left. 
    +s_t(\vx_t)-s_t(\vx_t|\vx_0)\right\rVert_2^2\right]\right],
\end{align}
where the unconditional score function $s_t(\vx)$ is estimated via an unconditional SGM $s(\vx_t,t;\vtheta)$. The CG-DLSM authors~\citep{chao2022denoising} prove that Eq.~\ref{eqn:dlsm} can calibrate the classifier to produce more accurate gradients $\nabla_x\log p(y| \vx)$.

Robust CGSGM~\citep{https://doi.org/10.48550/arxiv.2208.08664}, in contrast to CG-DLSM, does not regularize by modeling the unconditional distribution. Instead, robust CGSGM leverages existing techniques to improve the robustness of the classifier against \textit{adversarial} perturbations. 
Robust CGSGM applies a gradient-based adversarial attack to the $\vx_t$ generated during the diffusion process, and uses the resulting adversarial example to make the classifier more robust.

\begin{figure*}[t!]
\begin{minipage}[c]{1.0\textwidth}
\centering
\begin{subfigure}{0.6\linewidth}
\includegraphics[width=\linewidth]{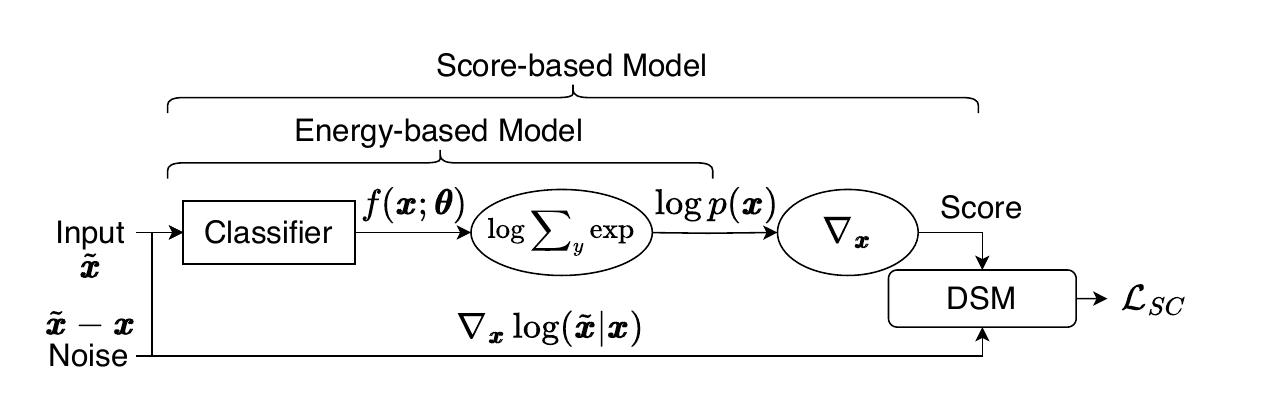}
\end{subfigure}
\vspace*{-2pt}
\caption{\textbf{Calculation of the proposed self-calibration loss.} First, perturbed sample $\tilde{x}$ is fed to the time-dependent classifier to obtain the output logits. Second, the logits are transformed into log-likelihood by applying $\log\sum_y\exp$. Third, we calculate its gradient w.r.t. the input $\tilde{x}$ to obtain the estimated score. Finally, denoising score matching is applied to obtain the proposed self-calibration loss. The proposed loss is used as an auxiliary loss to train the classifier.}
\label{fig:sc_procedure}
\end{minipage}
\end{figure*}

\subsection{Reinterpreting Classifiers as Energy-based models}
\label{sec:jem}
Our proposed methodology draws inspiration from JEM~\citep{Grathwohl2020Your}, which shows that reinterpreting classifiers as energy-based models (EBMs) and enforcing regularization with related objectives helps classifiers to capture more accurate probability distributions. EBMs are models that estimate energy functions $E(\vx)$ of distributions~\citep{lecun2006tutorial}, which satisfies $\log p(\vx)=-E(\vx)+\log\int_\vx\exp(E(\vx))d\vx$.
Given the logits of a classifier to be $f(\vx,y;\pmb{\phi})$, the estimated joint distribution can be written as $p(\vx,y; \pmb{\phi})=\frac{\exp(f(\vx,y;\pmb{\phi}))}{Z(\pmb{\phi})}$, where $\exp(\cdot)$ means exponential and $Z(\pmb{\phi})= \int_{\vx, y} \exp(f(\vx,y;\pmb{\phi}))\,d\vx\,dy$. After that, the energy function $E(\vx;\pmb{\phi})$ can be obtained by
\vspace{-0.1cm}
\begin{align}
    E(\vx;\pmb{\phi}) = -\log\Sigma_y\exp(f(\vx,y;\pmb{\phi}))
    \label{eqn:jem}
\end{align}
Then, losses used to train EBMs can be seamlessly leveraged in JEM to regularize the classifier, such as the typical EBM loss 
$\mathcal{L}_{\mathrm{EBM}}=\E_{p(\vx)}\left[-\log p(\vx; \pmb{\phi})\right]$. 
JEM uses MCMC sampling for computing the loss and is shown to result in a well-calibrated classifier in their empirical study. The original JEM work~\citep{Grathwohl2020Your} also reveals that classifiers can be used as a reasonable generative model, but its generation performance is knowingly worse than state-of-the-art SGMs.

\section{The Proposed Self-calibration Methodology}
\label{sec:method}

In this work, we consider CGSGMs under the diffusion generation process as discussed in Section~\ref{sec:back_sde}. Such CGSGMs require learning an unconditional SGM, which is assumed to be trained with denoising score matching~\citep[DSM;][]{Vincent2011Denoise} due to its close relationship with the diffusion process. Such CGSGMs also require a time-dependent classifier that models $p_t(y | \vx)$ instead of $p(y | \vx)$, which can be done by applying a time-generalized cross-entropy loss.

Section~\ref{sec:classguide} has illustrated the importance of regularizing the classifier to prevent it from misguiding the conditional generation process. One naive thought is to use established regularization techniques, such as label-smoothing and Jacobian regularization~\citep{hoffman2019robust}. Those regularization techniques that are less attached to the underlying distribution will be studied in Section~\ref{sec:experiments}. Our proposed regularization loss, inspired by the success of DLSM~\cite{chao2022denoising} and JEM~\citep{Grathwohl2020Your}, attempts to connect with the underlying distribution better.

\subsection{Formulation of self-calibration loss}
\label{sec:formulation}

We extend JEM~\citep{Grathwohl2020Your} to connect the time-dependent classifier to the underlying distribution. In particular, we reinterpret the classifier as a time-dependent EBM. The interpretation allows us to obtain a time-dependent version of $p_t(\vx)$ within the classifier, which can be used to obtain a classifier-\textit{internal} version of the score function. Then, instead of regularizing the classifier by the EBM loss $-\log p_t(\vx)$ like JEM, we propose to regularize by score function $\nabla_\vx \log p_t(\vx)$ instead.

Under the EBM interpretation, the energy function is
$E(\vx,t;\pmb{\phi}) = -\log\Sigma_y\exp(f(\vx,y,t;\pmb{\phi}))$, where $f(\vx,y,t;\pmb{\phi})$ is the output logits of the time-dependent classifier. Then, the internal time-dependent unconditional score function is
$s^c(\vx,t;\pmb{\phi})=\nabla_\vx \log\Sigma_y\exp(f(\vx,y,t;\pmb{\phi}))$, where $s^c$ is used instead of $s$ to indicate that the unconditional score is computed \textit{within} the classifier. Then, we adopt the standard DSM technique in Eq.~\ref{eqn:dsm} to ``train'' the internal score function, forcing it to follow its physical meaning during the diffusion process. The resulting self-calibration loss can then be defined as
\vspace{-0.1cm}
\begin{align}
\label{eqn:scloss}
    &\mathcal{L}_{\mathrm{SC}}(\pmb{\phi})\nonumber\\&=\E_t\left[\lambda(t)\E_{\vx_t,\vx_0}\left[\frac{1}{2}\left\lVert s^c(\vx_t,t;\pmb{\phi})-s_t(\vx_t|\vx_0)\right\rVert_2^2\right]\right],
\end{align}
where $\vx_t\sim p_t$, $\vx_0\sim p_0$, and $s_t(\vx_t|\vx_0)$ denotes the score function of the noise centered at $\vx_0$.

Fig.~\ref{fig:sc_procedure} summarizes the calculation of the proposed SC loss. Note that in practice, $t$ is uniformly sampled over $[0,T]$. After the self-calibration loss is obtained, it is mixed with the cross-entropy loss $\mathcal{L}_{\mathrm{CE}}$ to train the classifier. The total loss can be written as:
\begin{equation}
    \mathcal{L}_{\mathrm{CLS}}(\pmb{\phi})=\mathcal{L}_{\mathrm{CE}}(\pmb{\phi})+\lambda_{SC}\mathcal{L}_{\mathrm{SC}}(\pmb{\phi}),
    \label{eqn:classloss}
\end{equation}
where $\lambda_{SC}$ is a tunable hyper-parameter. The purpose of self-calibration is to make the classifier to more accurately estimate the score function of the underlying data distribution, implying that the underlying data distribution itself is also more accurately estimated. As a result, the gradients of the classifiers are more aligned with the ground truth. After self-calibration, the classifier is then used in CGSGM to guide an unconditional SGM for conditional generation. Note that since our approach regularizes the classifier during training while classifier gradient scaling (Eq.~\ref{eqn:cg_scale}) is done during sampling, we can easily combine the two techniques to enhance performance.

\begin{figure*}[t]
\begin{minipage}[c]{1.0\linewidth}
\centering
	\raisebox{-0.5\height}{\begin{subfigure}{0.1598\linewidth}
	\includegraphics[width=\linewidth, frame]{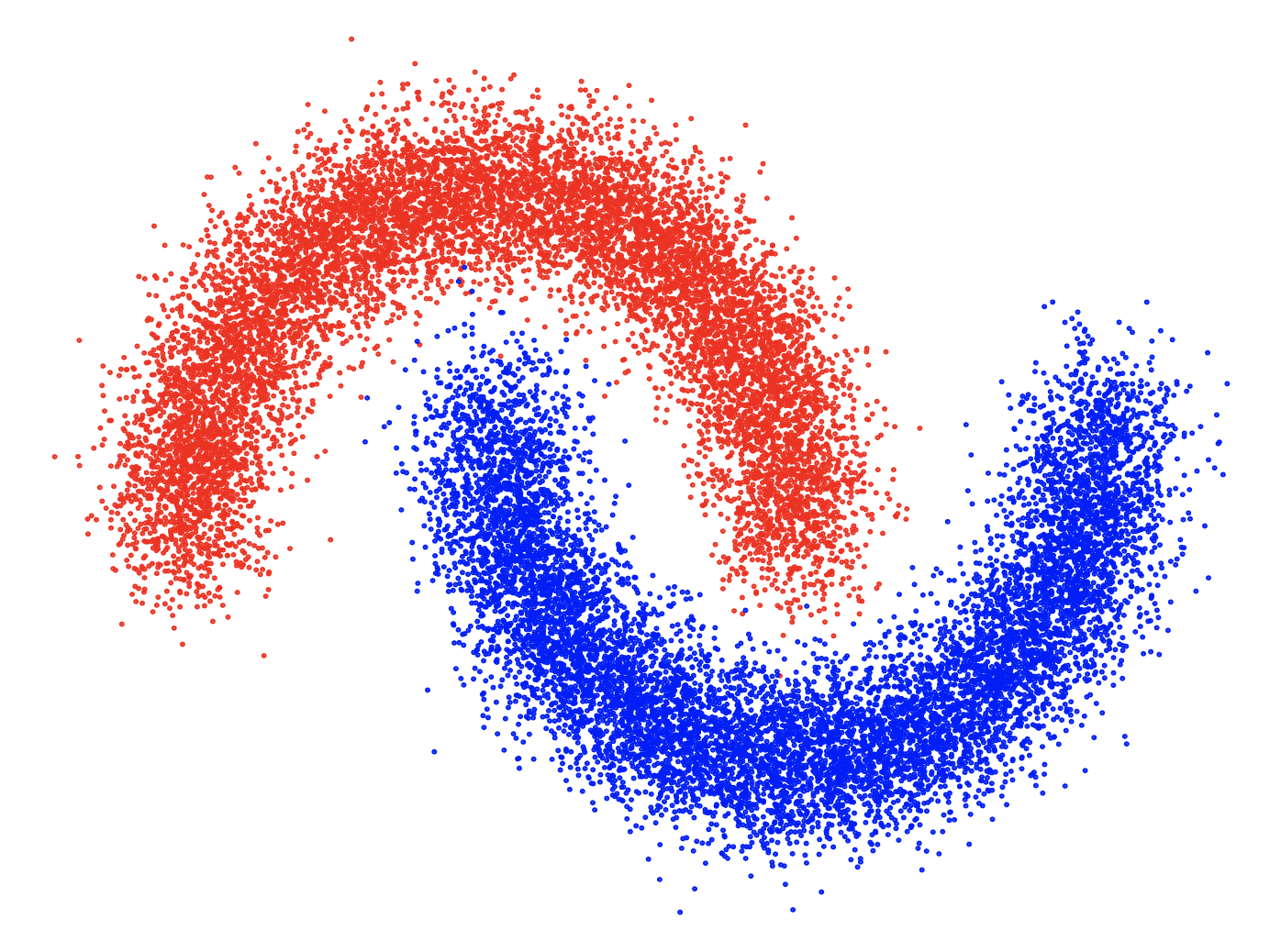}
        \subcaption{Toy data}
        \label{fig:toy_data}
	\end{subfigure}}
	\raisebox{-0.5\height}{\begin{subfigure}{0.1612\linewidth}
	\includegraphics[width=\linewidth,frame]{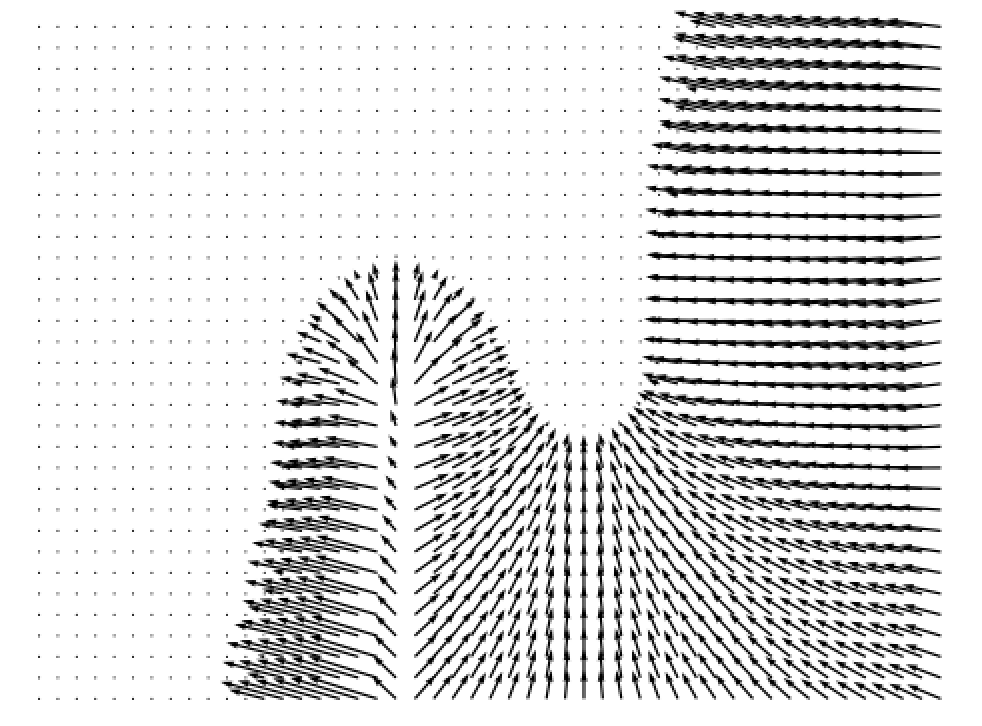}
	\includegraphics[width=\linewidth,frame]{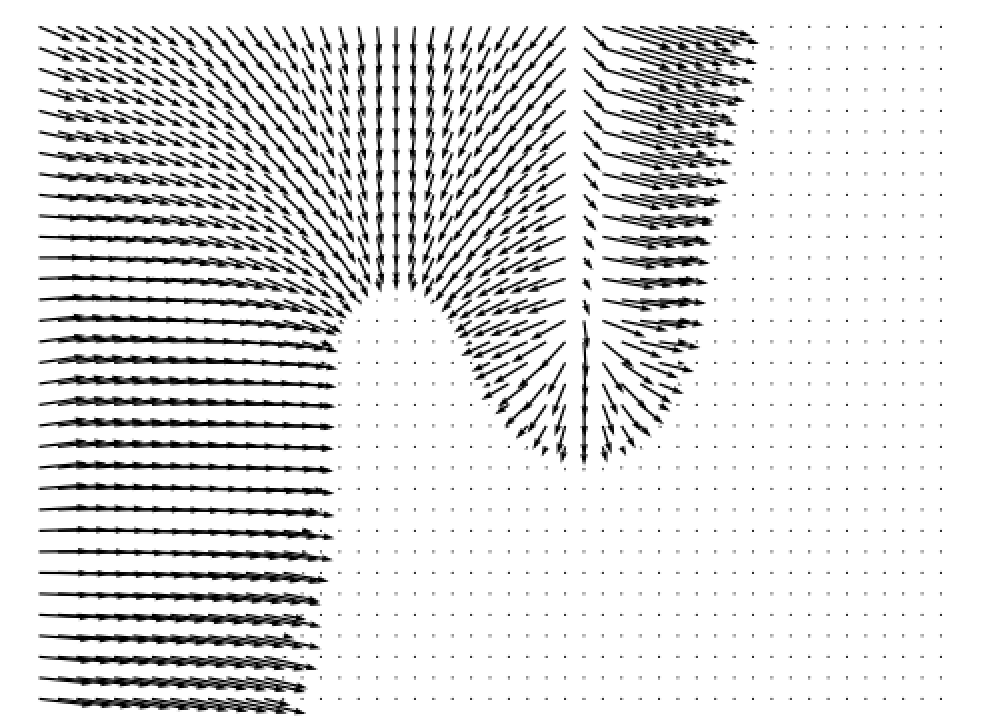}
	\subcaption{Ground truth}
        \label{fig:gt_grad}
	\end{subfigure}}
	\raisebox{-0.5\height}{\begin{subfigure}{0.161\linewidth}
	\includegraphics[width=\linewidth,frame]{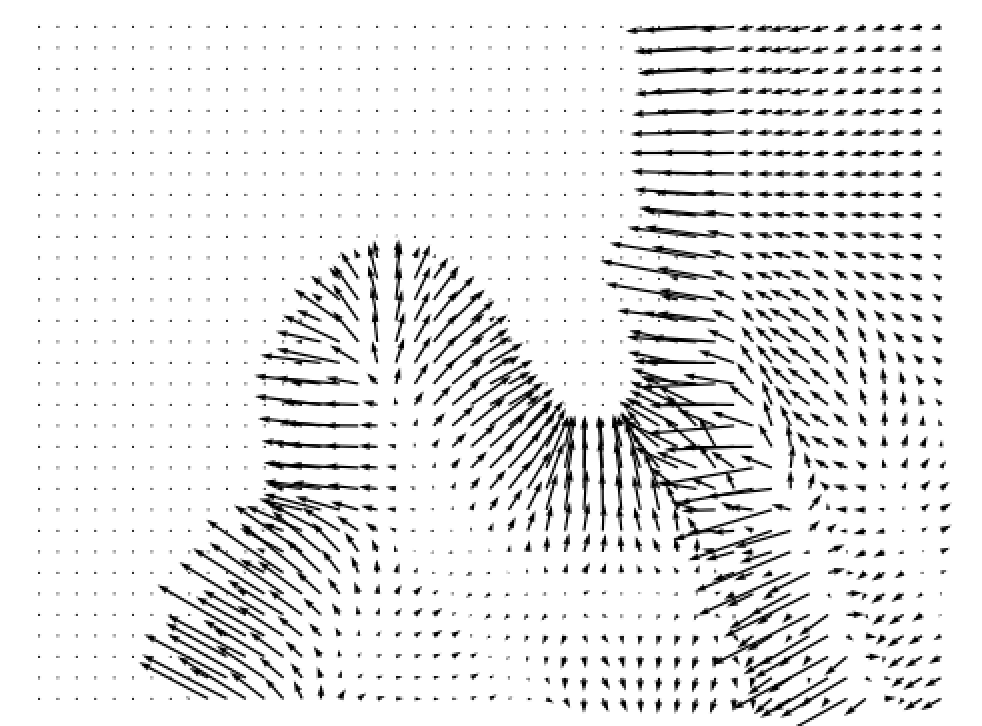}
	\includegraphics[width=\linewidth,frame]{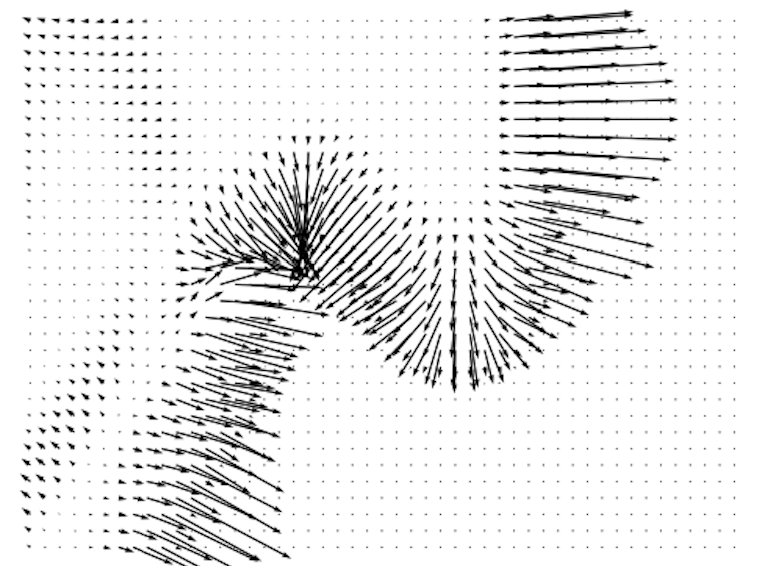}
	\subcaption{Vanilla}
        \label{fig:uncalib_grad}
	\end{subfigure}}
	\raisebox{-0.5\height}{\begin{subfigure}{0.1611\linewidth}
	\includegraphics[width=\linewidth,frame]{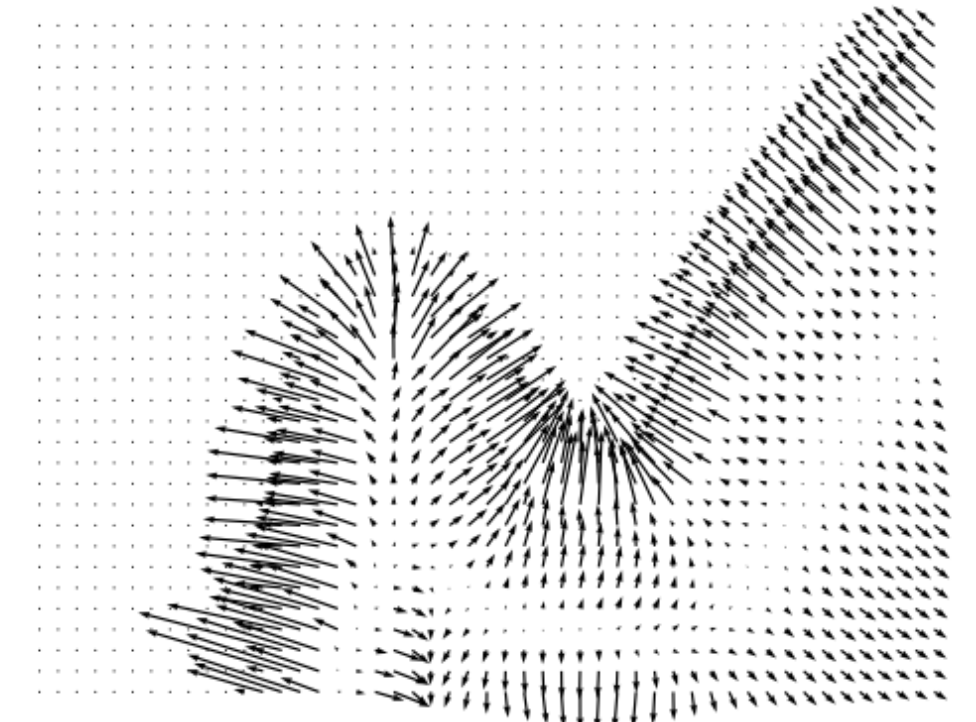}
	\includegraphics[width=\linewidth,frame]{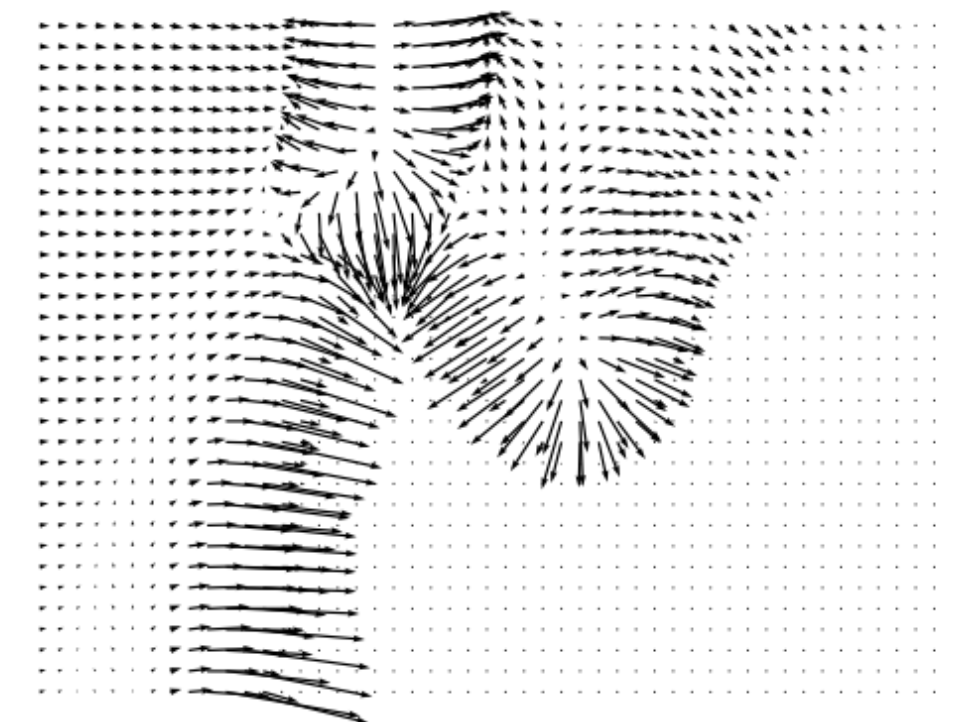}
	\subcaption{CG-DLSM}
        \label{fig:dlsm_grad}
	\end{subfigure}}
	\raisebox{-0.5\height}{\begin{subfigure}{0.1611\linewidth}
	\includegraphics[width=\linewidth,frame]{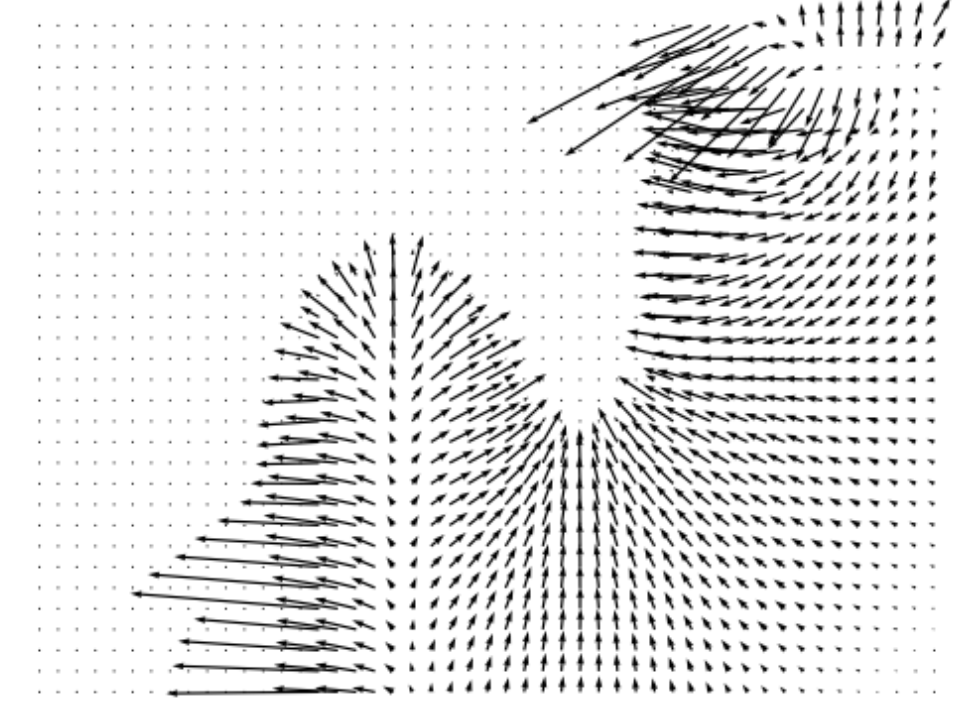}
	\includegraphics[width=\linewidth,frame]{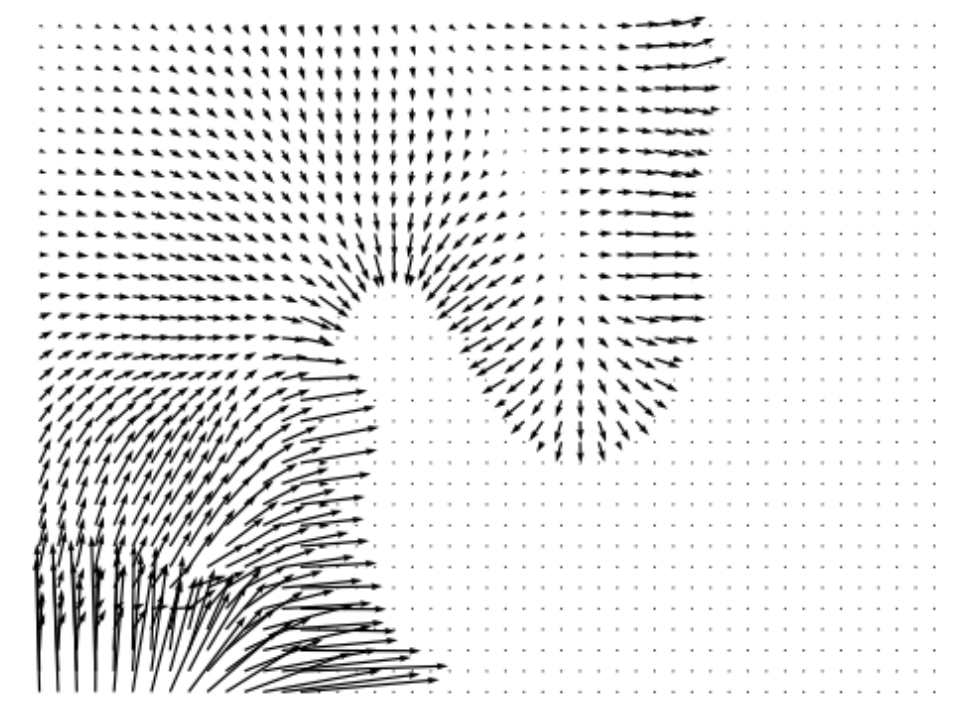}
	\subcaption{CG-JEM}
        \label{fig:jem_grad}
	\end{subfigure}}
	\raisebox{-0.5\height}{\begin{subfigure}{0.1607\linewidth}
	\includegraphics[width=0.994\linewidth,frame]{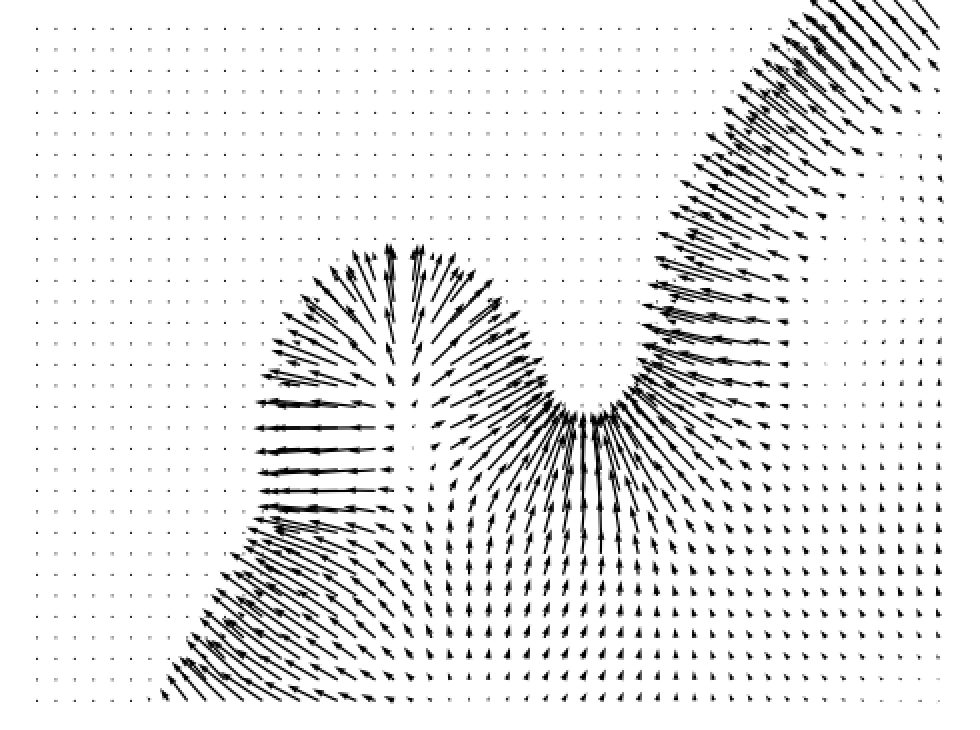}
	\includegraphics[width=\linewidth,frame]{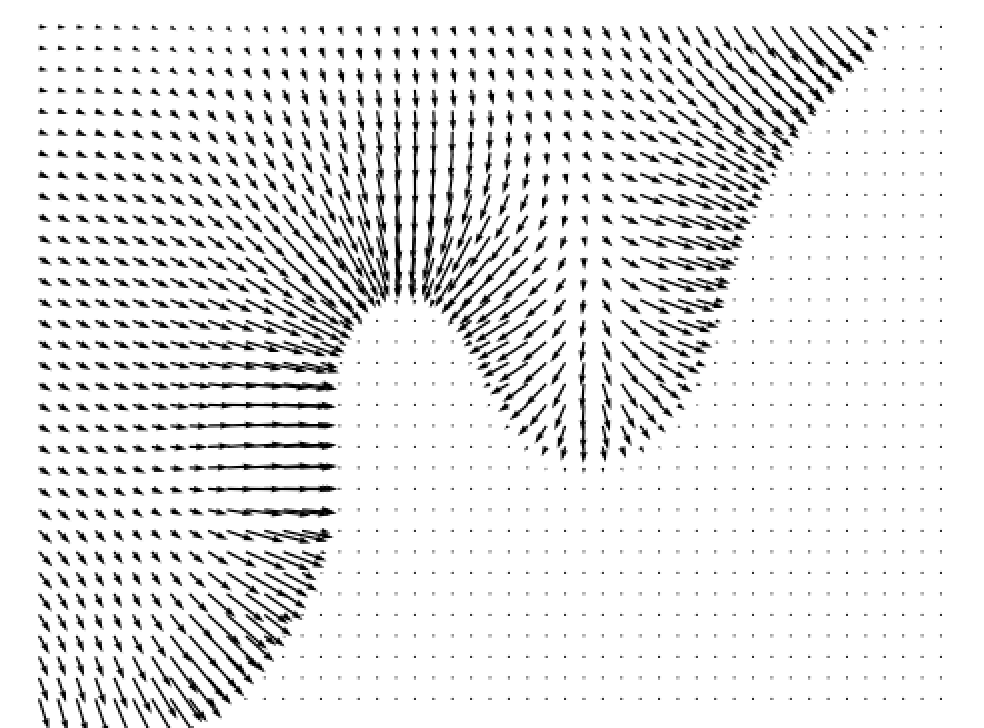}
	\subcaption{CG-SC (Ours)}
        \label{fig:calib_grad}
	\end{subfigure}}
\end{minipage}
\caption{\textbf{Gradients of classifiers $\nabla_\vx \log p(y|\vx)$ for toy dataset.} The upper row contains the gradients for class~1 (red), and the lower contains the gradients for class~2 (blue). (a)~Real data distribution. (b)~Ground truth classifier gradients. Gradients estimated by (c)~Vanilla CG, (d)~CG-DLSM, (e)~CG-JEM, and (f)~CG with proposed self-calibration. We observed that the gradients estimated by the vanilla classifier are highly inaccurate and fluctuate greatly. On the other hand, regularized classifiers produce gradients that are closer to the ground truth and contain much fewer fluctuations.}
\label{fig:toy_vec}
\end{figure*}

\subsection{Comparison with related regularization methods}
\label{sec:back_compare}
This section provides a comparative analysis of the regularization methods employed in DLSM~\citep{chao2022denoising}, robust classifier guidance~\citep{https://doi.org/10.48550/arxiv.2208.08664}, JEM~\cite{Grathwohl2020Your}, and our proposed self-calibration loss.

\paragraph{DLSM~\citep{chao2022denoising}} DLSM and our proposed method both regularize the classifier to align better with unconditional SGMs' view of the underlying distribution. DLSM achieves this by relying on the help of an \textit{external} trained SGM, whereas self-calibration regularizes the classifier by using a classifier-\textit{internal} SGM.
Furthermore, the design of DLSM loss can only utilize labeled data, while our method is able to make use of all data.

\paragraph{Robust CGSGM~\citep{https://doi.org/10.48550/arxiv.2208.08664}} Robust CGSGM proposes to regularize classifiers with gradient-based adversarial training without explicitly aligning with the distribution, in contrast to our self-calibration method, where direct calibration of the classifier-estimated score function is employed. Although adversarial robustness is correlated with more interpretable gradients~\citep{tsipras2018robustness}, EBMs~\citep{Zhu_2021_ICCV}, and generative modeling~\citep{Santurkar2019ImageSW}, the theoretical foundation for whether adversarially robust classifiers accurately estimate underlying distributions remains ambiguous. 

\paragraph{JEM~\citep{Grathwohl2020Your}} JEM interprets classifiers as unconditional EBMs, and self-calibration further extends the interpretation to unconditional SGMs. The training stage of EBM that incorporates MCMC sampling is known to be unstable and time-consuming, whereas self-calibration precludes the need for sampling during training and substantially improves both stability and efficiency. Even though one can mitigate the instability issue by increased sampling steps and additional hyperparameter tuning, doing so largely lengthens training times.

\paragraph{CG-JEM} In contrast to the previous paragraph, this paragraph discusses the incorporation of JEM into the time-dependent CGSGM framework. Coupling EBM training with additional training objectives is known to introduce increased instability, especially for time-dependent classifiers considering it is more difficult to generate meaningful time-dependent data through MCMC sampling. For example, in our naive implementation of time-dependent JEM, it either (1)~incurs high instability (loss diverges within $10,000$ steps in all $10$ runs; requires $4.23$s per step) or (2)~poses unaffordable resource consumption requirements (requires $20$s per step, approximately $50$ days in total). In comparison, our proposed method only requires $0.75$s per step.
While enhanced stability is observed after incorporating diffusion recovery likelihood~\citep{gao2021learning}, the time-costly nature of MCMC sampling still remains (requires $6.03$s per step).

\subsection{Self-calibration for Semi-supervised Learning}
In this work, we also explore the benefit of self-calibration loss in the semi-supervised setting, where only a small proportion of data are labeled. In the original classifier guidance, the classifiers are trained only on labeled data. The lack of labels in the semi-supervised setting constitutes a greater challenge to learning an unbiased classifier. With self-calibration, we better utilize the large amount of unlabeled data by calculating the SC loss with all data.

To incorporate the loss and utilize unlabeled data during training, we change the way $\mathcal{L}_{\mathrm{CLS}}$ is calculated from Eq.~\ref{eqn:classloss}. As illustrated in Fig.~\ref{fig:cg_illus}, the entire batch of data is used to calculate $\mathcal{L}_{\mathrm{SC}}$, whereas only the labeled data is used to calculate $\mathcal{L}_{\mathrm{CE}}$.
During training, we observe that when the majority is unlabeled data, the cross-entropy loss does not converge to a low-and-steady stage if the algorithm randomly samples from all training data. As this may be due to the low percentage of labeled data in each batch, we change the way we sample batches by always ensuring that exactly half of the data is labeled. Appendix~\ref{appendix:alg_sc} summarizes the semi-supervised training process of the classifier.

Note that even though the classifier is learning a time-generalized classification task, we can still make it perform as an ordinary classifier that classifies the unperturbed data by setting the input timestep $t=0$. This greatly facilitates the incorporation of common semi-supervised classification methods such as pseudo-labeling~\citep{Lee2013PseudoLabelT}, self-training, and noisy student~\citep{Xie_2020_CVPR}. Integrating semi-supervised classification methodologies is an interesting future research direction, and we reserve the detailed exploration of this topic for future studies.

\subsection{2D toy dataset}
\label{sec:toy}

Following DLSM~\citep{chao2022denoising}, we use a 2D toy dataset containing two classes to demonstrate the effects of self-calibration loss and visualize the training results. The data distribution is shown in Fig.~\ref{fig:toy_data}, where the two classes are shown in two different colors. After training the classifiers on the toy dataset with different methods, we plot the gradients $\nabla_\vx \log p(y|\vx)$ at minimum timestep $t=0$ estimated by the classifiers and compare them with the ground truth. Additional quantitative measurements of the toy dataset are included in Appendix~\ref{appendix:toy}.

Figure~\ref{fig:toy_vec} shows the ground truth classifier gradient and the gradients estimated by classifiers trained using different methods. Unregularized classifiers produce gradients that contain rapid changes in magnitude across the 2D space, with frequent fluctuations and mismatches with the ground truth. Such fluctuations can impede the convergence of the reverse diffusion process to a stable data point, leading SGMs to generate noisier samples. Moreover, the divergence from the ground truth gradient can misguide the SGM, leading to generation of samples from incorrect classes. Unregularized classifiers also tend to generate large gradients near the distribution borders and tiny gradients elsewhere. This implies that when the sampling process is heading toward the incorrect class, such classifiers are not able to ``guide'' the sampling process back towards the desired class. In comparison, the introduction of various regularization techniques such as DLSM, JEM, and the proposed self-calibration results in estimated gradients that are more stable, continuous across the 2D space, and better aligned with the ground truth. This stability brings about a smoother generation process and the production of higher-quality samples.

\section{Experiments}
\label{sec:experiments}
In this section, we test our methods on the CIFAR-10 and CIFAR-100 datasets. We demonstrate that our method improves generation both conditionally and unconditionally with different percentages of labeled data. Randomly selected images of CGSGM before and after self-calibration on CIFAR-10 are shown in Appendix~\ref{appendix:sample}. Additional experimental details are included in Appendix~\ref{appendix:exp_details}.

\subsection{Experimental Setup}
\label{sec:exp_setup}

\paragraph{Evaluation metrics} We adopted unconditional metrics FID~\citep{NIPS2017_8a1d6947}, density, and coverage~\citep{ferjad2020icml}. Density and coverage are designed to address the issues of precision and recall~\citep{NEURIPS2019_0234c510} to provide a more reliable interpretation of how well the generated data distribution resembles and covers the entirety of training data distribution. Besides unconditional generative performance, we evaluated the class-conditional performance of our methods using intra-FID, intra-density, and intra-coverage, which measures the average FID, density, and coverage for each class, respectively.

\begin{figure*}[t]
\centering
\begin{minipage}[c]{1.0\linewidth}
	\centering\begin{subfigure}{0.9\linewidth}
	\includegraphics[width=\linewidth]{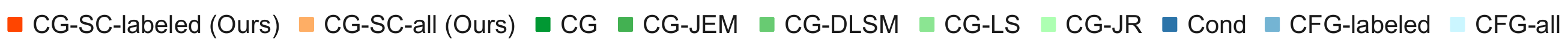}
         \end{subfigure}
\end{minipage}
\centering
\begin{minipage}[c]{1.0\linewidth}
\centering
        \begin{subfigure}{0.3\linewidth}
        \includegraphics[width=\linewidth]{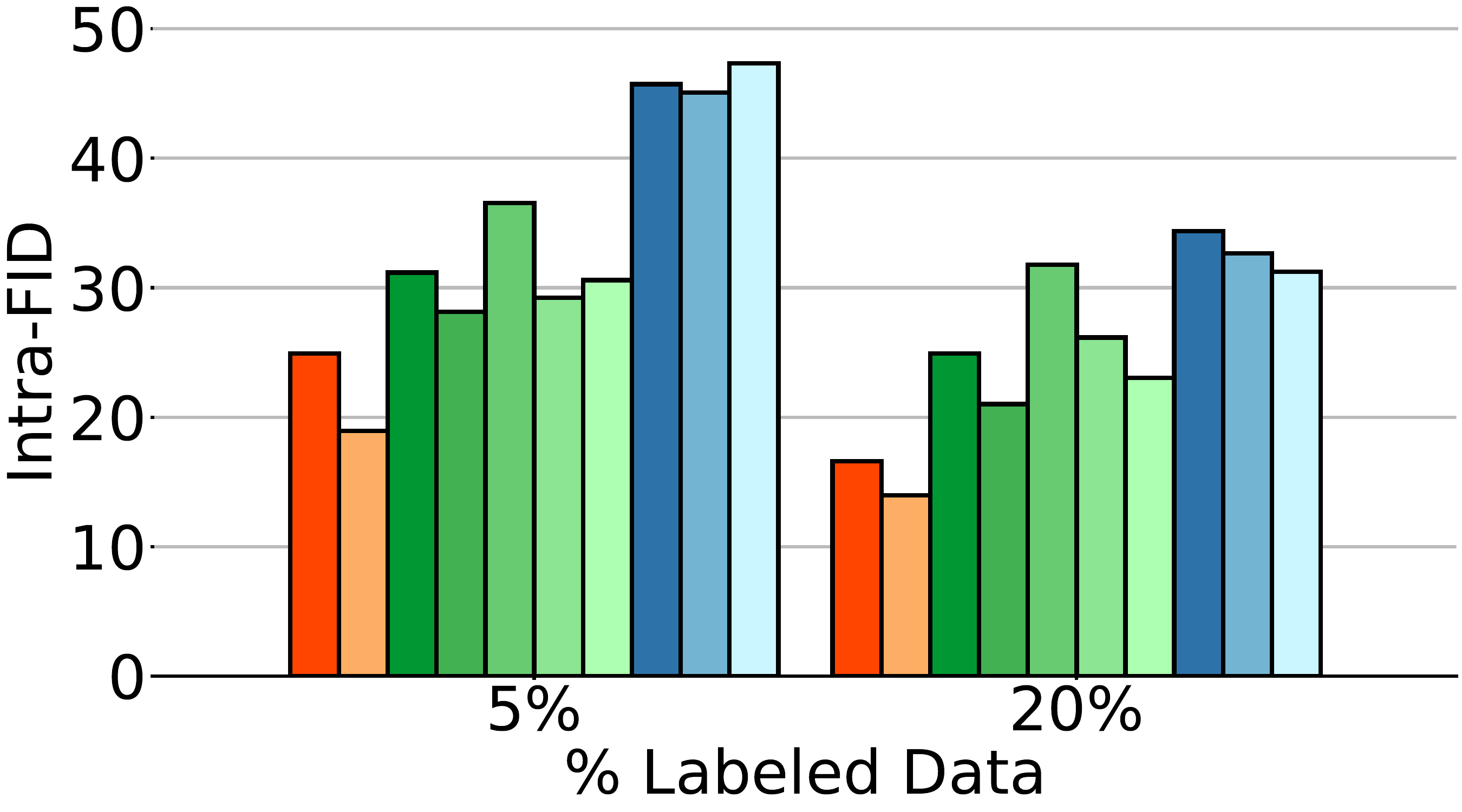}
        \subcaption{CIFAR-10: intra-FID ($\downarrow$)}
        \end{subfigure}
        \hspace{0.01\linewidth}
        \begin{subfigure}{0.3\linewidth}
        \includegraphics[width=\linewidth]{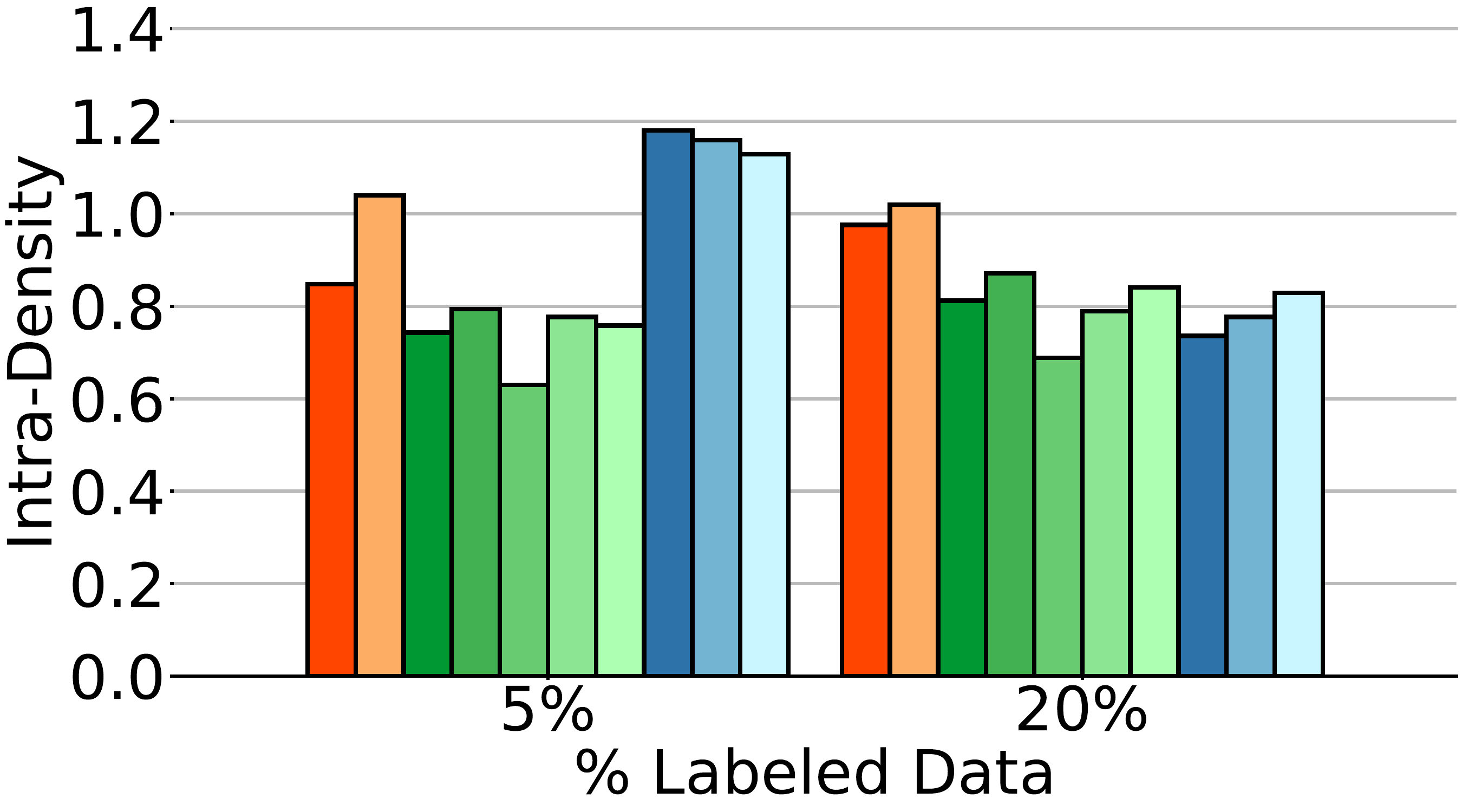}
        \subcaption{CIFAR-10: intra-Density ($\uparrow$)}
        \label{fig:semi_intra_density_cifar10}
        \end{subfigure}
        \hspace{0.01\linewidth}
        \begin{subfigure}{0.3\linewidth}
        \includegraphics[width=\linewidth]{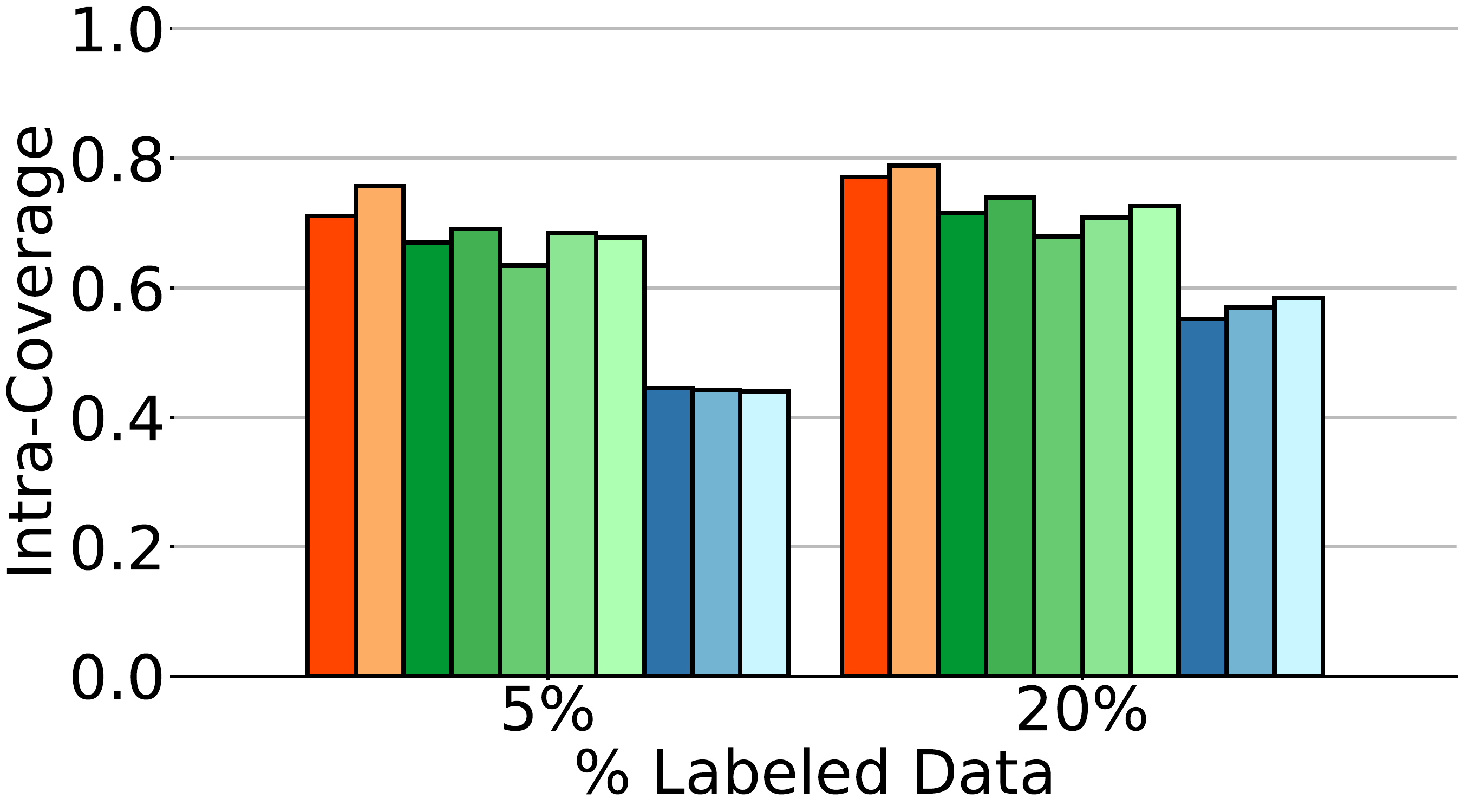}
        \subcaption{CIFAR-10: intra-Coverage ($\uparrow$)}
        \end{subfigure}
        
        \begin{subfigure}{0.3\linewidth}
        \includegraphics[width=\linewidth]{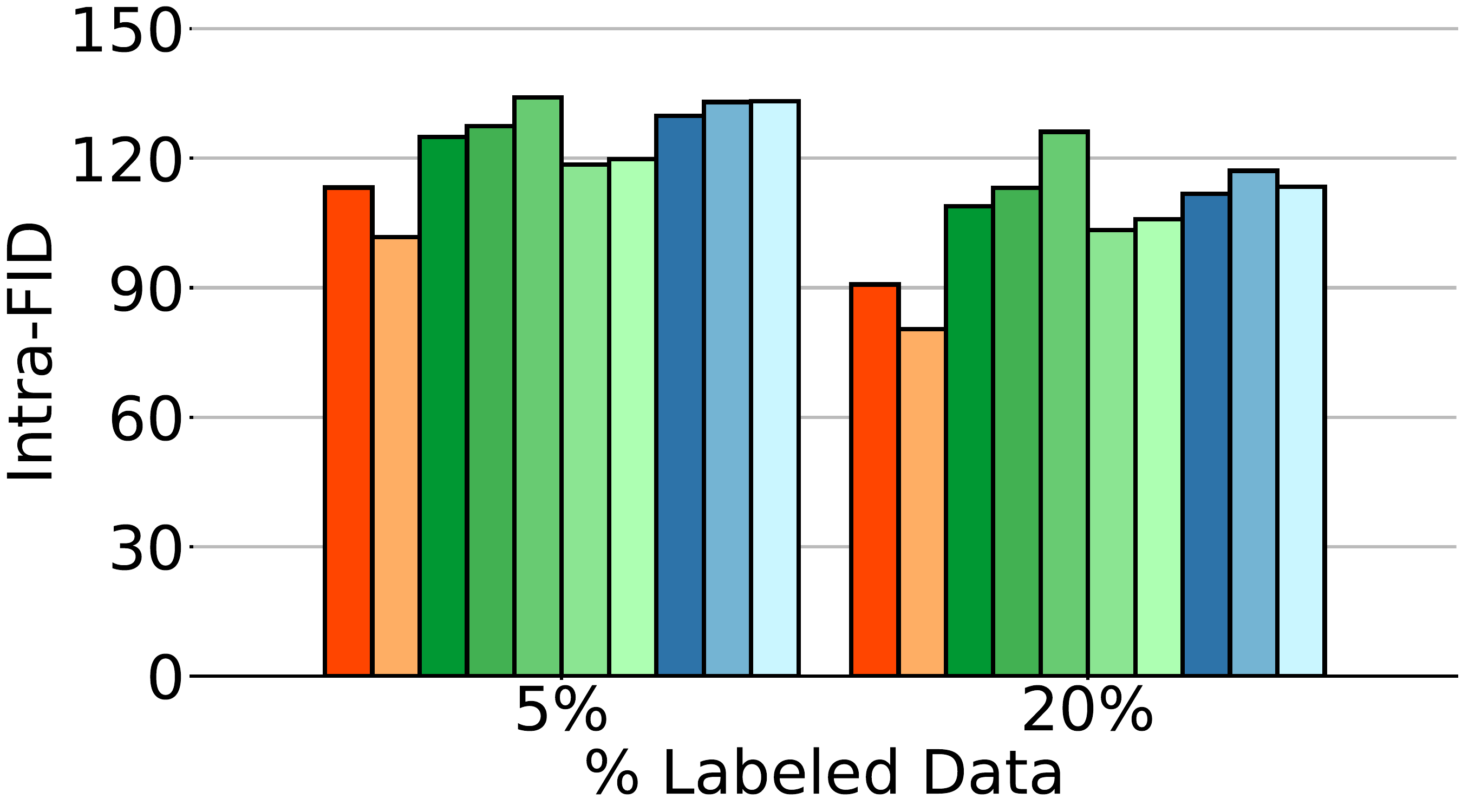}
        \subcaption{CIFAR-100: intra-FID ($\downarrow$)}
        \end{subfigure}
        \hspace{0.01\linewidth}
        \begin{subfigure}{0.3\linewidth}
        \includegraphics[width=\linewidth]{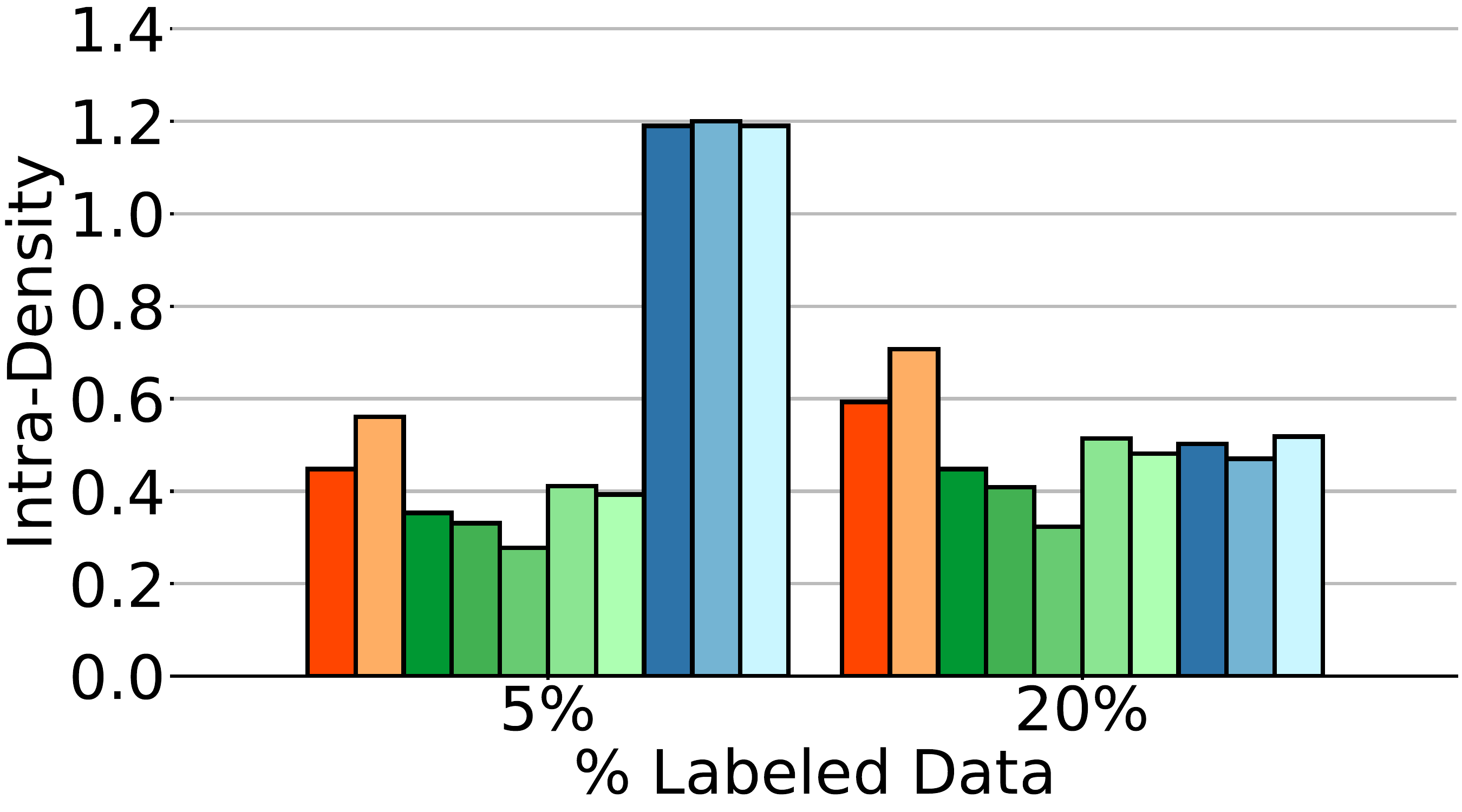}
        \subcaption{CIFAR-100: intra-Density ($\uparrow$)}
        \label{fig:semi_intra_density_cifar100}
        \end{subfigure}
        \hspace{0.01\linewidth}
        \begin{subfigure}{0.3\linewidth}
        \includegraphics[width=\linewidth]{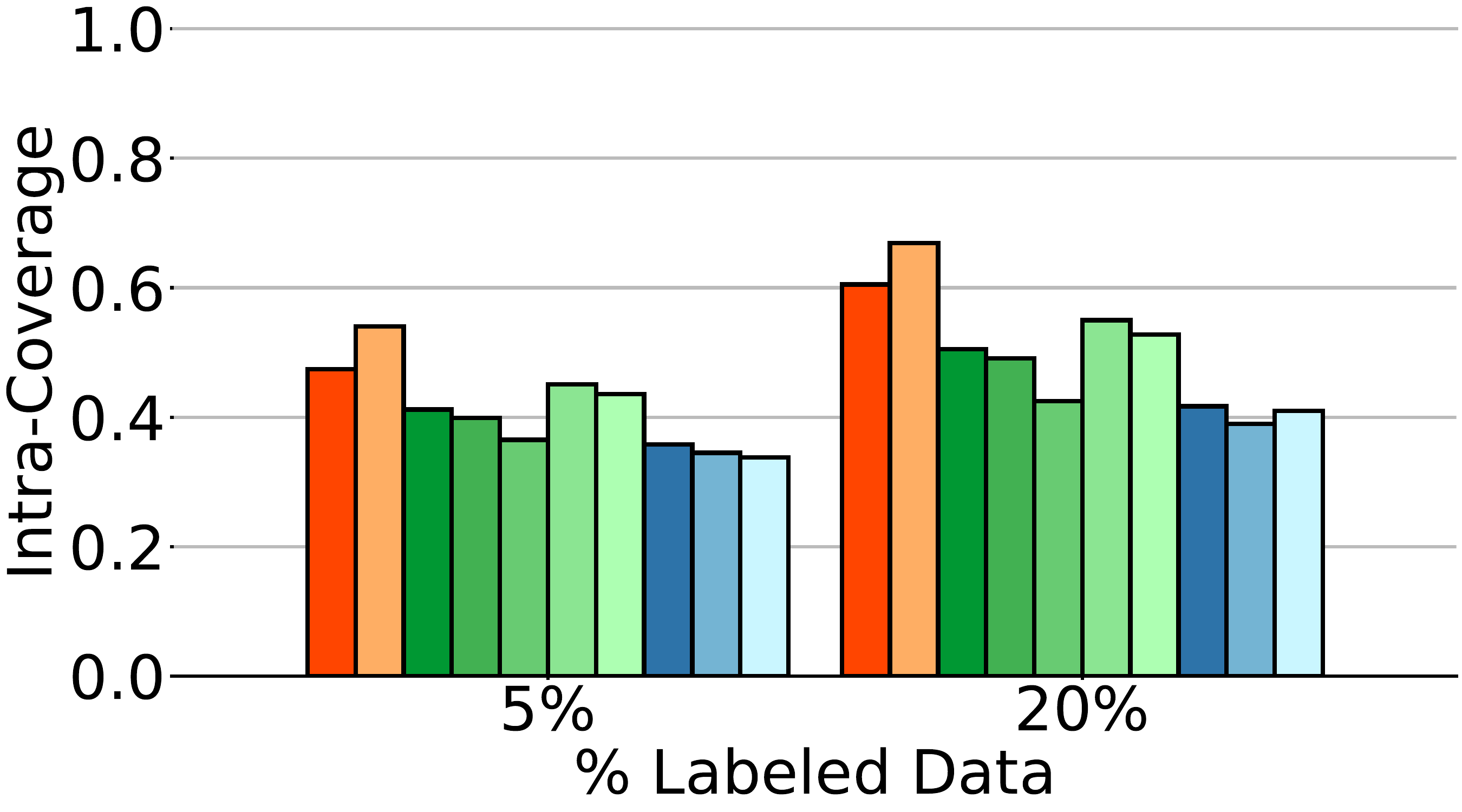}
        \subcaption{CIFAR-100: intra-Coverage ($\uparrow$)}
        \end{subfigure}
\end{minipage}
\vspace{-0.15cm}
\caption{\textbf{Results of class-conditional generation in semi-supervised settings.} In semi-supervised settings, although CFSGMs generate images with high intra-Density, which demonstrates they are able to generate images to the right class, the images only cover a small portion of the class-conditional distributions. This makes CGSGMs preferable as fewer labeled data does not cause the generation diversity of CGSGMs to decrease by much. Compared to other CGSGMs, the proposed self-calibration consistently achieves the best conditional generative performance.}
\vspace{-0.15cm}
\label{fig:semi_result}
\end{figure*}

\paragraph{Baseline methods} \textbf{CG}: vanilla classifier guidance; \textbf{CG-JEM}: classifier guidance with JEM loss; \textbf{CG-DLSM}: classifier guidance with DLSM loss~\citep{chao2022denoising}; \textbf{CG-LS}: classifier guidance with label smoothing; \textbf{CG-JR}: classifier guidance with Jacobian regularization~\citep{hoffman2019robust}; \textbf{Cond}: conditional SGMs by conditional normalization~\citep{dhariwal2021diffusion}; \textbf{CFG-labeled}: CFG~\citep{ho2021classifierfree} using only labeled data; \textbf{CFG-all}: CFG using all data to train the unconditional part of the model.

\subsection{Results}
\label{sec:semi_result}
Table~\ref{table:semi_supervised} and Fig.~\ref{fig:semi_result} present the performance of all methods when applied to various percentages of labeled data.
\textbf{CG-SC-labeled} implies that self-calibration is applied only on labeled data whereas \textbf{CG-SC-all} implies that self-calibration is applied on all data.

\paragraph{Classifier-Free SGMs (CFSGMs)}
The first observation from Fig.~\ref{fig:semi_result} is that CFSGMs, including Cond, CFG-labeled, and CFG-all, consistently excel in the intra-Density metric, demonstrating their ability to generate class-conditional images with high accuracy.
However, in Table~\ref{table:semi_supervised} and Fig.~\ref{fig:semi_result}, we can see that when performing class-conditional generation tasks with few labeled data, there is a significant performance drop in terms of Coverage, intra-FID, and intra-Coverage for CFSGMs.
CFSGMs, while generating high-quality images with high accuracy, tend to lack diversity when working with fewer labeled data.
This occurs mainly due to the lack of sufficient labeled data in the training phase, causing them to generate samples that closely mirror the distribution of only the labeled data, as opposed to that of all data. This inability to generate data that covers the full variability of training data deteriorated Coverage and intra-Coverage, which also caused CFSGMs to perform poorly in terms of intra-FID, despite being able to generate class-conditional images with high accuracy.

\paragraph{Classifier-Guided SGMs (CGSGMs)}
By leveraging both labeled and unlabeled data during training in semi-supervised settings, all CGSGMs are able to perform better in semi-supervised class-conditional generation compared to CFSGMs.
Besides better performance, the diversity measures, which contain coverage, and intra-Coverage, of CGSGMs are also more consistent across various percentages of labeled data (Table~\ref{table:semi_supervised}). This demonstrates that having less labeled data does not reduce the diversity of generated images by much for CGSGMs, which is preferable over CFSGMs as the generation diversity of CFSGMs is greatly deteriorated by the reduction of labeled data.

\begin{table*}[t]
\begin{minipage}[c]{1.0\linewidth}
\setlength{\tabcolsep}{1.4pt}\scriptsize\centering
 \caption{\textbf{Sample quality comparison of all methods with various percentages of labeled data.} \textbf{Bold}: best performance among all methods; \underline{underlined}: best performance among CG-based methods. Den and Cov stands for Density and Coverage, respectively. CGSGMs demonstrate superior performance compared to CFSGMs in semi-supervised settings. Furthermore, the proposed CG-SC-labeled and CG-SC-all outperform other CGSGMs in semi-supervised settings.}
\label{table:semi_supervised}
\begin{subtable}[b]{\linewidth}\centering
\vspace{-0.2cm}
\subcaption{Results of semi-supervised settings on CIFAR-10 dataset}
\label{table:semi_supervised_cifar10}
\begin{tabular}{lcccccccccccc}
\toprule
 & \multicolumn{4}{c}{5\% labeled data} & \multicolumn{4}{c}{20\% labeled data} & \multicolumn{4}{c}{100\% labeled data} \\ \cmidrule(lr){2-5} \cmidrule(lr){6-9} \cmidrule(lr){10-13}
Method & intra-FID ($\downarrow$) & FID ($\downarrow$) & Den ($\uparrow$) & Cov ($\uparrow$) & intra-FID ($\downarrow$) & FID ($\downarrow$) & Den ($\uparrow$) & Cov ($\uparrow$) & intra-FID ($\downarrow$) & FID ($\downarrow$) & Den ($\uparrow$) & Cov ($\uparrow$) \\ \midrule
CG-SC-labeled (Ours) & 24.93 & 2.84 & 1.083 & 0.816 & 16.62 & 2.75 & \underline{\textbf{1.097}} & \underline{\textbf{0.823}} & \underline{11.70} & 2.23 & \underline{1.029} & \underline{0.817} \\
CG-SC-all (Ours) & \underline{\textbf{18.95}} & 2.72 & \underline{\textbf{1.191}} & \underline{\textbf{0.822}} & \underline{\textbf{13.97}} & 2.63 & 1.090 & 0.821 & \underline{11.70} & 2.23 & \underline{1.029} & \underline{0.817} \\
\midrule
 CG & 31.17 & 2.61 & 1.047 & 0.815 & 24.94 & 3.09 & 1.004 & 0.806 & 18.99 & 2.48 & 0.979 & 0.812 \\
CG-JEM & 28.13 & 2.70 & 1.079 & 0.817 & 21.02 & 2.75 & 1.050 & 0.817 & 23.39 & 2.16 & 0.980 & 0.813 \\
CG-DLSM & 36.55 & \underline{\textbf{2.18}} & 0.992 & 0.816 & 31.78 & \underline{\textbf{2.10}} & 0.975 & 0.812 & 21.59 & 2.36 & 0.943 & 0.803 \\
CG-LS & 29.24 & 2.62 & 1.068 & 0.821 & 26.15 & 4.18 & 0.971 & 0.796 & 18.10 & \underline{2.15} & 1.005 & 0.818 \\
CG-JR & 30.59 & 2.80 & 1.066 & 0.817 & 23.03 & 2.49 & 1.056 & 0.822 & 17.24 & 2.17 & 1.007 & 0.815 \\
Cond & 45.73 & 15.57 & 1.174 & 0.459 & 34.36 & 19.77 & 0.775 & 0.567 & \textbf{10.29} & \textbf{2.13} & 1.050 & 0.831 \\
CFG-labeled & 45.07 & 15.31 & 1.169 & 0.457 & 32.66 & 18.48 & 0.832 & 0.589 & 10.58 & 2.28 & \textbf{1.101} & \textbf{0.838} \\
CFG-all & 47.33 & 16.57 & 1.144 & 0.452 & 31.24 & 17.37 & 0.863 & 0.601 & 10.58 & 2.28 & \textbf{1.101} & \textbf{0.838} \\
\bottomrule
\end{tabular}
\end{subtable}

\begin{subtable}[b]{\linewidth}\centering
\setlength{\tabcolsep}{3.0pt}
\subcaption{Results of semi-supervised settings on CIFAR-100 dataset}
\label{table:semi_supervised_cifar100}
\begin{tabular}{lcccccccccccc}
\toprule
 & \multicolumn{4}{c}{5\% labeled data} & \multicolumn{4}{c}{20\% labeled data} & \multicolumn{4}{c}{100\% labeled data} \\ \cmidrule(lr){2-5} \cmidrule(lr){6-9} \cmidrule(lr){10-13}
Method & intra-FID ($\downarrow$) & FID ($\downarrow$) & Den ($\uparrow$) & Cov ($\uparrow$) & intra-FID ($\downarrow$) & FID ($\downarrow$) & Den ($\uparrow$) & Cov ($\uparrow$) & intra-FID ($\downarrow$) & FID ($\downarrow$) & Den ($\uparrow$) & Cov ($\uparrow$) \\ \midrule
CG-SC-labeled (Ours) & 113.21 & 4.80 & 0.927 & 0.756 & 90.76 & 3.74 & 0.928 & 0.766 & \underline{79.57} & 3.70 & 0.848 & 0.749 \\
CG-SC-all (Ours) & \underline{\textbf{101.75}} & 4.31 & \underline{0.968} & \underline{\textbf{0.770}} & \underline{\textbf{80.42}} & \underline{\textbf{3.60}} & \underline{\textbf{0.941}} & 0.775 & \underline{79.57} & 3.70 & 0.848 & 0.749 \\
\midrule
 CG & 124.92 & 5.24 & 0.891 & 0.745 & 108.86 & 4.10 & 0.904 & 0.759 & 98.72 & 3.83 & 0.869 & 0.761 \\
CG-JEM & 127.47 & 6.01 & 0.831 & 0.723 & 113.16 & 5.08 & 0.864 & 0.754 & 106.24 & 5.40 & 0.818 & 0.730 \\
CG-DLSM & 134.11 & 4.46 & 0.862 & 0.749 & 126.12 & 7.24 & 0.802 & 0.718 & 102.85 & 3.85 & 0.847 & 0.753 \\
CG-LS & 118.52 & \underline{\textbf{4.18}} & 0.933 & 0.762 & 103.39 & 3.70 & 0.963 & \underline{\textbf{0.778}} & 98.53 & \underline{3.39} & \underline{0.907} & \underline{0.774} \\
CG-JR & 119.78 & 4.64 & 0.926 & 0.759 & 105.91 & 3.92 & \underline{\textbf{0.941}} & 0.771 & 100.34 & 3.50 & 0.892 & 0.768 \\
Cond & 129.82 & 10.58 & 1.210 & 0.404 & 111.73 & 29.45 & 0.622 & 0.451 & 64.77 & 3.02 & 0.962 & 0.806 \\
CFG-labeled & 133.03 & 11.25 & \textbf{1.220} & 0.385 & 117.09 & 32.68 & 0.611 & 0.424 & \textbf{63.03} & \textbf{2.60} & \textbf{1.040} & \textbf{0.829} \\
CFG-all & 133.18 & 10.68 & 1.210 & 0.380 & 113.38 & 30.84 & 0.612 & 0.437 & \textbf{63.03} & \textbf{2.60} & \textbf{1.040} & \textbf{0.829}
\\ \bottomrule
\end{tabular}
\end{subtable}
\end{minipage}
\end{table*}

\paragraph{Regularized CGSGMs vs Vanilla CGSGM}
Basic regularization methods like label-smoothing and Jacobian regularization~\citep{hoffman2019robust} show consistent but marginal improvement over vanilla CGSGM. This points out that although these methods mitigate overfitting on training data, the constraints they enforce do not align with SGMs, limiting the benefit of such methods.
CG-DLSM~\citep{chao2022denoising}, on the other hand, achieves great unconditional generation performance in all settings. However, its class-conditional performance suffers from a significant performance drop in semi-supervised settings due to its low generation accuracy, which is evident in Fig.~\ref{fig:semi_intra_density_cifar10} and Fig.~\ref{fig:semi_intra_density_cifar100}.
We can also see in Fig.~\ref{fig:semi_result} and Table~\ref{table:semi_supervised} that by incorporating the proposed self-calibration, all conditional metrics improved substantially.
Notably, CG-SC-all (Ours) consistently achieves the best conditional performance among all CGSGMs in semi-supervised settings. On average, CG-SC-all improves intra-FID by 10.16 and 23.59 over CG on CIFAR-10 and CIFAR-100, respectively.
These results demonstrate that self-calibration enables CGSGMs to estimate the class-conditional distributions more accurately, even when labeled data is limited.

\paragraph{Leverage unlabeled data for semi-supervised settings}
Intuitively, we expect incorporating unlabeled data into the computation of SC loss to enhance the quality of conditional generation. As the proportion of unlabeled data increases, we expect this benefit of leveraging unlabeled data to become more significant.
As our experimental results indicate in Fig.~\ref{fig:semi_result} and Table~\ref{table:semi_supervised}, the utilization of unlabeled data significantly improves performance.
Notably, with only 5\% of the data labeled on CIFAR-10, the performance of CG-SC-all evaluated with intra-FID and intra-Density are improved by 5.98 and 0.156 over CG-SC-labeled and 12.22 and 0.296 over the original CG.
The results confirm that as the percentage of labeled data decreases, the benefit of utilizing unlabeled data becomes increasingly more profound.

\section{Conclusion}

We tackle the overfitting issue for the classifier within CGSGMs from a novel perspective: self-calibration. The proposed self-calibration method leverages EBM interpretation like JEM to reveal that the classifier is internally an unconditional score estimator and design a loss with the DSM technique to calibrate the internal estimation. This self-calibration loss regularizes the classifier directly towards better scores without relying on an external score estimator. We demonstrate three immediate benefits of the proposed self-calibrating CGSGM approach. Using a standard synthetic dataset, we show that the scores computed using this approach are indeed closer to the ground-truth scores. Second, across all percentages of labeled data, the proposed approach outperforms the existing CGSGM. Lastly, our empirical study justifies that when compared to other conditional SGMs, the proposed approach consistently achieves the best intra-FID in the focused semi-supervised settings by seamlessly leveraging the power of unlabeled data, highlighting the rich potential of our approach.

We have presented compelling evidence supporting the superiority of CGSGM using standard CIFAR-10 and CIFAR-100 datasets. A natural question arises regarding whether this superiority can be extended to higher-resolution datasets such as ImageNet. We conjecture that achieving this extension is highly non-trivial, as previous studies~\citep{song2021scorebased, song2021maximum, chao2022denoising, Ma2022AccelerateSGM} have not successfully developed a satisfactory \textit{unconditional} SGM on ImageNet when the resolution exceeds 32 by 32. CGSGM on high-resolution ImageNet, which naturally inherits the technical limitation from unconditional SGMs, thus remains challenging to build, despite our extensive exploration through hyperparameter tuning and analysis. In addition to responsibly reporting this limitation to the community's attention, two immediate future directions emerge: the first involves conducting foundational research to achieve competent generation for unconditional SGMs on ImageNet, and the second explores applying the self-regularization concept to other types of conditional generative models.

\section{Impact Statements}
We expect our work to assist the development of real-world applications in scenarios where labeled data is scarce. However, as advancements in conditional generation methodologies introduce improved controllability over synthetic data, the proposed method may make intentional or unintentional misuse of generative models easier. For instance, fraudulent content could be generated, or privacy leaks could be caused by instructing models to synthesize data closely mirroring specific training samples.

\bibliographystyle{plainnat}  
\bibliography{references}  
\newpage
\appendix

\section{Supplementary Experimental Results on CIFAR-10}
\label{appendix:class_result}

\subsection{Additional semi-supervised learning settings}
In Section~\ref{sec:experiments}, we discussed the generative performance using 5\%, 20\%, and 100\% labeled data from CIFAR-10. In this section, we provide further results for scenarios where 40\%, 60\%, and 80\% of the data is labeled. Besides the evaluation metrics used in the main paper, Inception Score (IS)~\citep{NIPS2016_8a3363ab} is also provided. Note that results for CG-JEM with 40\%, 60\%, and 80\% labeled data are not included due to limited computational resources.
\begin{table}[h!]
\begin{minipage}[c]{1.0\linewidth}
\setlength{\tabcolsep}{1.4pt}\scriptsize\centering
\caption{Sample quality comparison of all methods. \textbf{Bold}: best performance among all methods; \underline{underlined}: best performance among CG-based methods.}
\label{table:semi_supervised_additional_cifar10}
\begin{subtable}[b]{\textwidth}\centering
\vspace{-0.3cm}
\subcaption{intra-FID, FID, and IS of the CIFAR-10 dataset}
\begin{tabular}{lccccccccc}
\toprule
 & \multicolumn{3}{c}{5\% labeled data} & \multicolumn{3}{c}{20\% labeled data} & \multicolumn{3}{c}{100\% labeled data} \\ \cmidrule(lr){2-4} \cmidrule(lr){5-7} \cmidrule(lr){8-10}
Method & intra-FID ($\downarrow$) & FID ($\downarrow$) & IS ($\uparrow$) & intra-FID ($\downarrow$) & FID ($\downarrow$) & IS ($\uparrow$) & intra-FID ($\downarrow$) & Acc ($\uparrow$) & IS ($\uparrow$) \\ \midrule
 CG-SC-labeled (Ours) & 24.93 & 2.84 & 9.78 & 16.62 &
                        2.75 & 9.83 & \underline{11.70} &
                        2.23 & 9.82 \\
 CG-SC-all (Ours) & \underline{\textbf{18.95}} & 2.72 & \underline{9.95} & \underline{\textbf{13.97}} &
                    2.63 & 9.94 & \underline{11.70} &
                    2.23 & 9.82 \\ \midrule
 CG & 31.17 & 2.61 & 9.98 & 24.94 &
              3.09 & 9.92 & 18.99 &
              2.48 &  9.88 \\
CG-JEM & 28.13 & 2.70 & 9.92 & 21.02 & 2.75 & 10.10 & 23.39 & 2.16 & 9.83 \\
 CG-DLSM & 36.55
    & \underline{\textbf{2.18}} & 9.76 & 31.78
    & \underline{\textbf{2.10}} & 9.91 & 21.59 & 2.36 & 9.92 \\
 CG-LS & 29.24
    & 2.62 & 9.92 & 26.15
    & 4.18 & 9.98 & 18.10 & \underline{2.15} & \underline{9.98} \\
 CG-JR & 30.59
    & 2.80 & 9.84 & 23.03
    & 2.49 & \underline{\textbf{10.04}} & 17.24 & 2.17 & 9.89 \\
 Cond & 45.73 & 15.57 & 9.87 & 34.36 &
           19.77 & 8.82 & \textbf{10.29} &
           \textbf{2.13} & \textbf{10.06} \\
 CFG-labeled & 45.07
    & 15.31 & \textbf{10.20} & 32.66
    & 18.48 & 8.93 & 10.58 & 2.28 & 10.05 \\
 CFG-all & 47.33
    & 16.57 & 9.89 & 31.24
    & 17.37 & 9.15 & 10.58 & 2.28 & 10.05 \\
\bottomrule
\toprule
 & \multicolumn{3}{c}{40\% labeled data} & \multicolumn{3}{c}{60\% labeled data} & \multicolumn{3}{c}{80\% labeled data} \\ \cmidrule(lr){2-4} \cmidrule(lr){5-7} \cmidrule(lr){8-10}
Method & intra-FID ($\downarrow$) & FID ($\downarrow$) & IS ($\uparrow$) & intra-FID ($\downarrow$) & FID ($\downarrow$) & IS ($\uparrow$) & intra-FID ($\downarrow$) & Acc ($\uparrow$) & IS ($\uparrow$) \\ \midrule
 CG-SC-labeled (Ours) & \underline{\textbf{12.08}} & 2.78 & 10.00 & \underline{11.65} & 
                        2.37 & 9.91 & \underline{11.86} &
                        2.24 & 9.78 \\
 CG-SC-all (Ours) & 12.67 & 2.72 & \underline{\textbf{10.04}} & 12.22 & 
                    2.42 & 9.95 & 12.47 &
                    2.25 & 9.83 \\ \midrule
 CG & 18.31 & 2.42 & 9.95 & 16.94 & 
      2.35 & 10.03 & 20.15 &
      3.30 &  9.76 \\
 CG-DLSM & 29.33
    & 2.35 & 9.85 & 23.52
    & \underline{\textbf{2.15}} & 9.83 & 21.76 & 2.30 & \underline{9.96} \\
 CG-LS & 17.89
    & \underline{\textbf{2.32}} & 9.95 & 17.72
    & 2.27 & 9.91 & 22.30 & 2.40 & 9.84 \\
 CG-JR & 18.63
    & 2.43 & 10.01 & 19.05
    & 2.25 & \underline{10.06} & 18.36 & \underline{\textbf{2.15}} & 9.90 \\
 Cond & 13.65 & 4.36 & 9.94 & \textbf{10.93} &
           2.55 & 10.00 & \textbf{10.61} &
           2.37 & 10.03 \\
 CFG-labeled & 13.93
            & 4.59 & 9.84 & 11.28 & 2.73 & \textbf{10.12} & 10.75 &
           2.48 & \textbf{10.09} \\
 CFG-all & 13.43 & 4.30 & 9.98 & 11.38 &
           2.83 & 10.05 & 10.94 & 
           2.50 & 10.03 \\
\bottomrule
\end{tabular}
\end{subtable}
\begin{subtable}[b]{\textwidth}\centering
\subcaption{Density, Coverage, intra-Density, and intra-Coverage of the CIFAR-10 dataset}
\begin{tabular}{lcccccccccccc}
\toprule
 & \multicolumn{4}{c}{5\% labeled data} & \multicolumn{4}{c}{20\% labeled data} & \multicolumn{4}{c}{100\% labeled data} \\ \cmidrule(lr){2-5} \cmidrule(lr){6-9} \cmidrule(lr){10-13}
Method & Den ($\uparrow$) & Cov ($\uparrow$) & intra-Den ($\uparrow$) & intra-Cov ($\uparrow$) & Den ($\uparrow$) & Cov ($\uparrow$) & intra-Den ($\uparrow$) & intra-Cov ($\uparrow$) & Den ($\uparrow$) & Cov ($\uparrow$) & intra-Den ($\uparrow$) & intra-Cov ($\uparrow$) \\ \midrule
CG-SC-labeled (Ours) & 1.083 & 0.816 & 0.848 & 0.711 & \underline{\textbf{1.097}} & \underline{\textbf{0.823}} & 0.976 & 0.771 & \underline{1.029} & 0.817 & \underline{0.992} & \underline{0.803} \\
CG-SC-all (Ours) & \underline{\textbf{1.191}} & \underline{\textbf{0.822}} & \underline{1.040} & \underline{\textbf{0.757}} & 1.090 & 0.821 & \underline{\textbf{1.020}} & \underline{\textbf{0.789}} & \underline{1.029} & 0.817 & \underline{0.992} & \underline{0.803} \\ \midrule
CG & 1.047 & 0.815 & 0.743 & 0.670 & 1.004 & 0.806 & 0.812 & 0.715 & 0.979 & 0.812 & 0.878 & 0.769 \\
CG-JEM & 1.079 & 0.817 & 0.794 & 0.691 & 1.050 & 0.817 & 0.871 & 0.739 & 0.980 & 0.813 & 0.781 & 0.723 \\
CG-DLSM & 0.992 & 0.816 & 0.630 & 0.634 & 0.975 & 0.812 & 0.688 & 0.680 & 0.943 & 0.803 & 0.779 & 0.734 \\
CG-LS & 1.068 & 0.821 & 0.777 & 0.685 & 0.971 & 0.796 & 0.789 & 0.708 & 1.005 & \underline{0.818} & 0.861 & 0.767 \\
CG-JR & 1.066 & 0.817 & 0.758 & 0.677 & 1.056 & 0.822 & 0.841 & 0.727 & 1.007 & 0.815 & 0.881 & 0.766 \\
Cond & 1.174 & 0.459 & \textbf{1.180} & 0.445 & 0.775 & 0.567 & 0.736 & 0.552 & 1.050 & 0.831 & 1.040 & 0.827 \\
CFG-labeled & 1.169 & 0.457 & 1.159 & 0.442 & 0.832 & 0.589 & 0.777 & 0.569 & \textbf{1.101} & \textbf{0.838} & \textbf{1.092} & \textbf{0.831} \\
CFG-all & 1.144 & 0.452 & 1.129 & 0.440 & 0.863 & 0.601 & 0.829 & 0.585 & \textbf{1.101} & \textbf{0.838} & \textbf{1.092} & \textbf{0.831} \\
\bottomrule
\toprule
 & \multicolumn{4}{c}{40\% labeled data} & \multicolumn{4}{c}{60\% labeled data} & \multicolumn{4}{c}{80\% labeled data} \\ \cmidrule(lr){2-5} \cmidrule(lr){6-9} \cmidrule(lr){10-13}
Method & Den ($\uparrow$) & Cov ($\uparrow$) & intra-Den ($\uparrow$) & intra-Cov ($\uparrow$) & Den ($\uparrow$) & Cov ($\uparrow$) & intra-Den ($\uparrow$) & intra-Cov ($\uparrow$) & Den ($\uparrow$) & Cov ($\uparrow$) & intra-Den ($\uparrow$) & intra-Cov ($\uparrow$) \\ \midrule
CG-SC-labeled (Ours) & \underline{1.080} & \underline{0.820} & \underline{1.040} & 0.810 & 1.000 & 0.810 & 0.962 & 0.796 & 1.030 & 0.820 & 0.992 & 0.803 \\
CG-SC-all (Ours) & 1.070 & \underline{0.820} & 1.030 & 0.804 & 1.000 & 0.810 & 0.947 & 0.795 & 1.030 & 0.820 & 0.992 & 0.803 \\ \midrule
CG & 1.030 & \underline{0.820} & 0.909 & 0.769 & 0.920 & 0.800 & 0.817 & 0.750 & 0.980 & 0.810 & 0.878 & 0.769 \\
CG-DLSM & 0.980 & 0.810 & 0.782 & 0.732 & 0.940 & 0.810 & 0.781 & 0.737 & 0.940 & 0.800 & 0.779 & 0.734 \\
CG-LS & 1.030 & \underline{0.820} & 0.894 & 0.768 & 0.970 & 0.810 & 0.788 & 0.737 & 1.010 & 0.820 & 0.861 & 0.767 \\
CG-JR & 1.020 & \underline{0.820} & 0.858 & 0.752 & 0.990 & 0.810 & 0.853 & 0.760 & 1.010 & 0.820 & 0.881 & 0.766 \\
Cond & 1.050 & \underline{0.820} & \underline{1.040} & \underline{0.816} & \underline{1.050} & \underline{0.830} & \underline{1.040} & \underline{0.821} & \underline{1.050} & \underline{0.830} & \underline{1.040} & \underline{0.827} \\
CFG-labeled & \textbf{1.100} & \textbf{0.830} & \textbf{1.089} & \textbf{0.822} & \textbf{1.120} & \textbf{0.840} & \textbf{1.111} & \textbf{0.834} & \textbf{1.100} & \textbf{0.840} & \textbf{1.092} & \textbf{0.831} \\
CFG-all & \textbf{1.100} & \textbf{0.830} & \textbf{1.089} & 0.820 & 1.110 & \underline{0.830} & 1.103 & 0.830 & \textbf{1.100} & \textbf{0.840} & \textbf{1.092} & \textbf{0.831} \\
\bottomrule
\end{tabular}
\end{subtable}
\end{minipage}
\end{table}

Table~\ref{table:semi_supervised_additional_cifar10} presents the results, further confirming the observations made in Section~\ref{sec:semi_result}. The CFSGMs consistently exhibit high generation accuracy but suffer from significant performance drops as the labeled data percentage decreases. Conversely, the CGSGMs maintain stable performance across various settings. Furthermore, our proposed CG-SC consistently outperforms other CG-based methodologies in terms of intra-FID and generation accuracy.

\subsection{Evaluation of expected calibration error}
Beyond the generative performance metrics, we present the Expected Calibration Error (ECE) to assess the calibration of classifiers regarding accurate probability estimation. ECE serves as a metric that evaluates the alignment of a classifier's confidence with its prediction accuracy.
The classifier's confidence is defined as:
\begin{align*}
    \mathrm{conf}(\vx)=\max_y p(y|\vx)=\max_y \frac{\exp(f(\vx,y;\pmb{\phi}))}{\sum_{y'}\exp(f(\vx,y';\pmb{\phi}))},
\end{align*}
where $f(\vx,y;\pmb{\phi})$ is the classifier's logits. We then divide the classifier's predictions based on confidence into several buckets. The average absolute difference between the confidence and prediction accuracy is calculated for each bucket. Then, given a labeled test set $D_{t}=\{(\vx_{m}, y_{m})\}_{m=1}^M$, ECE is defined as:
\begin{align*}
    \mathrm{ECE} = \sum_{i=1}^N\frac{\lvert B_i\rvert}{\lvert D_t\rvert}\cdot\left\lvert \mathrm{Acc}(B_i) - \frac{1}{\lvert B_i\rvert}\sum_{\vx\in B_i} \mathrm{conf}(\vx)\right\rvert,
\end{align*}
where $N$ is the number of buckets, $B_i=\left\{\vx\vert \mathrm{conf}(\vx)\in [\frac{i-1}{N},\frac{i}{N})\right\}$, $\mathrm{Acc}(B_i)$ is the averaged classification accuracy of $B_i$, and $\frac{1}{\lvert B_i\rvert}\sum_{\vx\in B_i}\mathrm{conf}(\vx)$ is the averaged confidence of $B_i$.

\begin{table}[h]
\begin{minipage}[c]{1.0\linewidth}
\scriptsize\centering
\caption{Expected calibration error ($\downarrow$) of all methods with various percentages of labeled data}
\label{table:ECE}
\begin{tabular}{lcccccc}
\toprule
Method & 5\% & 20\% & 40\% & 60\% & 80\% & 100\% \\ \midrule
 CG-SC-labeled (Ours) & 0.369 & 0.316 & \textbf{0.087} & \textbf{0.057} & \textbf{0.063} & \textbf{0.031}  \\
 CG-SC-all (Ours) & 0.210 & \textbf{0.243} & 0.102 & 0.109 & 0.111 & \textbf{0.031} \\ \midrule
 CG & 0.460 & 0.330 & 0.269 & 0.190 & 0.163 & 0.112 \\
 CG-DLSM & 0.468 & 0.343 & 0.307 & 0.237 & 0.180 & 0.183 \\
 CG-LS & \textbf{0.194} & 0.257 & 0.101 & 0.063 & 0.081 & 0.050 \\
 CG-JR & 0.407 & 0.348 & 0.279 & 0.225 & 0.183 & 0.173 \\
\bottomrule
\end{tabular}
\end{minipage}
\end{table}

We follow the setup in~\citet{Grathwohl2020Your}, setting $N=20$ for our calculations. The results are shown in Table~\ref{table:ECE}, illustrating the ECE values for all CG-based methods across various percentages of labeled data. Our observations underscore that the self-calibration method consistently enhances classifier ECE in comparison to the vanilla CG and delivers the most superior ECE in most cases. This validates our claim that self-calibrated classifiers offer a more accurate probability estimation.

\section{More detailed introduction on score-based generative modeling through SDE}
\label{appendix:more_on_sde}

\subsection{Learning the score function}
When learning the score function, the goal is to choose the best function from a family of
functions $\{s(\vx; \vtheta)\}_{\vtheta}$, such as deep learning models
parameterized by $\vtheta$, to approximate the score function $\nabla_\vx \log
p(\vx)$ of interest. Learning is based on data $\{\vx_n\}_{n=1}^N$
assumed to be sampled from $p(\vx)$. It has been shown that this
can be achieved by optimizing the in-sample version of the following
score-matching loss over $\theta$:
\begin{align*}
\mathcal{L}_{\mathrm{SM}}=\E_{p(\vx)}\left[tr(\nabla_\vx s(\vx;\vtheta))+\frac{1}{2}\left\lVert s(\vx;\vtheta)\right\rVert^2_2\right],
\end{align*}
where $tr(\cdot)$ denotes the trace of a matrix and $\nabla_\vx
s(\vx;\vtheta)=\nabla^2_\vx \log p(x)$ is the Hessian matrix of log-likelihood
$\log p(\vx)$. Calculating the score-matching loss requires  $O(d)$
computation passes for $\vx \in \sR^d$, which makes the optimization process
computationally prohibitive on high-dimensional data.

Several attempts~\citep{NIPS2010_6f3e29a3, Martens2012EstimatingTH,
Vincent2011Denoise, song2019sliced} have been made to address these computational
challenges by approximating or transforming score matching into equivalent
objectives. One current standard approach is called denoise score matching
(DSM)~\citep{Vincent2011Denoise}, which instead learns the score function of a
noise-perturbed data distribution $q(\tilde{\vx})$. DSM typically
assumes that $q(\tilde{\vx})$ comes from the original distribution $p(\vx)$
injected with a pre-specified noise $q(\tilde{\vx}|\vx)$. It has been
proved~\citep{Vincent2011Denoise} that the score function can be learned by
minimizing the in-sample version of
\begin{align*}
\E_{q(\tilde{\vx}|\vx)p(\vx)}\left[\frac{1}{2}\left\lVert s(\tilde{\vx};\vtheta) - \nabla_{\tilde{\vx}}\log q(\tilde{\vx}|\vx)\right\rVert_2^2\right],
\end{align*}
where $\nabla_{\tilde{\vx}}\log q(\tilde{\vx}|\vx)$ is the score function of
the noise distribution centered at $\vx$. DSM is generally more efficient than
the original score matching and is scalable to high-dimensional data as it
replaces the heavy computation on the Hessian matrix with simple perturbations
that can be efficiently computed from data.

\subsection{Generation from the score function by diffusion}
\label{appendix:generate_from_sde}
Assume that we seek to sample from some unknown target distribution $p(\vx) = p_0(\vx)$, and the distribution can be diffused to a known prior distribution $p_T(\vx)$ through a Markov chain that is described with
a stochastic differential equation (SDE)~\citep{song2021scorebased}:
$d\vx=f(\vx,t)dt+g(t)d\vw$,
where the Markov chain is computed for $0 \leq t < T$ using the drift function $f(\vx,t)$ that describes the overall movement and the dispersion function $g(t)$ that describes how the noise $\vw$ from a standard Wiener process enters the system. 

To sample from $p(\vx) = p_0(\vx)$, \citet{song2021scorebased} proposes to reverse the SDE from $p_T(\vx)$ to $p_0(\vx)$, which turns out to operate with another SDE (Eq.~\ref{eqn:revsde}). Given the score function $s(\vx, t)$, the diffusion process in Eq.~\ref{eqn:revsde} can then be used to take any data sampled from the known $p_T(\vx)$ to a sample from the unknown $p(\vx) = p_0(\vx)$.

The time-dependent score function $s(\vx, t; \vtheta)$ can be learned by minimizing a time-generalized (in-sample) version of the DSM loss because the diffusion process can be viewed as one particular way of injecting noise. The extended DSM loss is defined as
\begin{align*}
    \mathcal{L}_{DSM}(\vtheta)=\E_t\left[\lambda(t)\E_{\vx_t,\vx_0}\left[\frac{1}{2}\left\lVert s(\vx_t,t;\vtheta) - s_t(\vx_t|\vx_0)\right\rVert_2^2\right]\right],
\end{align*}
where $t$ is selected uniformly between $0$ and~$T$, $\vx_t\sim p_t(\vx)$, $\vx_0\sim p_0(\vx)$, $s_t(\vx_t|\vx_0)$ denotes the score function of $p_t(\vx_t | \vx_0)$, and $\lambda(t)$ is a weighting function that balances the loss of different timesteps.

\section{Training algorithm for semi-supervised self-calibrating classifier}
\label{appendix:alg_sc}
\begin{algorithm*}[h]
   \caption{Semi-supervised classifier training with self-calibration loss}
   \label{alg:semi_sc}
\begin{algorithmic}
   \STATE {\bfseries Input:} Labeled data $D_l$, unlabeled data $D_u$
   \STATE Initialize the time-dependent classifier $f(\vx,y,t;\pmb{\phi})$ randomly
   \REPEAT
   \STATE Sample data $(\vx_l, y_l)\sim D_l$, $\vx_u\sim D_u$
   \STATE Sample timesteps $t_l,t_u\sim \mathrm{Uniform}(1,T)$
   \STATE Obtain perturbed data $\tilde{\vx_l}\sim p_{t_l}(\vx|\vx_l),\tilde{\vx_u}\sim p_{t_u}(\vx|\vx_u)$
   \STATE Calculate $\mathcal{L}_{CE}=\mathrm{CrossEntropy}(f(\vx_l,y,t;\pmb{\phi}), y_l)$
   \STATE Calculate $\mathcal{L}_{SC}=\E_{(\vx,t)\in \{(\vx_l,t_l),(\vx_u,t_u)\}}\left[\frac{1}{2}\lambda(t)\lVert \nabla_\vx\log\Sigma_y\exp(f(\vx,y,t;\pmb{\phi}))-s_t(\vx_t|\vx_0)\rVert^2_2\right]$
   \STATE Take gradient step on $\mathcal{L}_{CLS}=\mathcal{L}_{CE}+\mathcal{L}_{SC}$
   \UNTIL{converged}
\end{algorithmic}
\end{algorithm*}

\section{Additional experimental details}
\label{appendix:exp_details}
We followed NCSN++~\citep{song2021scorebased} to implement the unconditional score estimation model. We also adapted the encoder part of NCSN++ as the classifier used in CGSGM~\citep{dhariwal2021diffusion} and its variants, e.g., CG-DLSM or the proposed CG-SC. For the sampling method, we used Predictor-Corrector (PC) samplers~\citep{song2021scorebased} with 1000 sampling steps. The SDE was selected as the VE-SDE framework proposed by~\citet{song2021scorebased}. The hyper-parameter introduced in Eq.~\ref{eqn:classloss} is tuned between $\{10,1,0.1,0.01\}$ for fully-supervised settings, and selected to be $1$ in semi-supervised settings due to limited computational resources. The scaling factor $\lambda_{CG}$ introduced in Eq.~\ref{eqn:cg_scale} is tuned within $\{0.5, 0.8, 1.0, 1.2, 1.5, 2.0, 2.5\}$ to obtain the best intra-FID. A similar scaling factor $\lambda_{CFG}$ for classifier-free SGMs is tuned within $\{0, 0.1, 0.2, 0.4\}$ to obtain the best intra-FID. The balancing factors of the DLSM loss and the Jacobian regularization loss are selected to be $1$ and $0.01$, respectively, as suggested in the original papers. The smoothing factor of label-smoothing is tuned between $\{0.1, 0.05\}$ for the better intra-FID.

\section{Quantitative measurements of toy dataset}
\label{appendix:toy}
\begin{table}[h]
\begin{minipage}[c]{1.0\linewidth}
\small\centering
\caption{Mean squared error (MSE) and cosine similarity (CS) of all CGSGM methods tested on the toy dataset}
\label{table:toy}
\begin{tabular}{cccc}
\toprule
Method & Gradient MSE ($\downarrow$) & Gradient CS ($\uparrow$) & Cond-Score CS ($\uparrow$)\\
\midrule
CG & 8.7664 & 0.3265 & 0.9175 \\
CG + scaling & 8.1916 & 0.3348 & 0.9447 \\
CG-SC & 7.1558 & 0.5667 & 0.9454 \\
CG-SC + scaling & \textbf{5.6376} & 0.5758 & 0.9689 \\
CG-DLSM & 8.1183 & 0.4450 & 0.9316 \\
CG-DLSM + scaling & 8.0671 & 0.4450 & 0.9328 \\
CG-JEM & 8.5577 & 0.6422 & 0.9670\\
CG-JEM + scaling & 8.5577 & \textbf{0.6429} & \textbf{0.9709} \\
\bottomrule
\end{tabular}
\end{minipage}
\end{table}

Table~\ref{table:toy} shows the quantitative measurements of the methods on the toy dataset. First, we compared the gradients $\nabla_\vx \log p(y|\vx)$ estimated by the classifiers with the ground truth by calculating the mean squared error (first column) and cosine similarity (second column). The results were calculated by averaging over all $(x,y)\in\{(x,y)|x\in \{-12,-11.5,\ldots,11.5,12\}, y\in\{-8,-7.5,\ldots,7.5,8.0\}\}$. We observe that after self-calibration, the mean squared error of the estimated gradients is $18\%$ lower; tuning the scaling factor further improves this to $36\%$. This improvement after scaling implies that the direction of gradients better aligns with the ground truth, and scaling further reduces the mismatch between the magnitude of the classifier and the ground truth. In terms of cosine similarity, self-calibration grants the classifiers an improvement of $42\%$. The numerical results agree with our previous observation that after self-calibration, classifiers better align with the ground truth in terms of both direction and magnitude.

Then, we add the unconditional score of the training data distribution to the classifier gradients to calculate the conditional scores and compare the results with the ground truth. The resulting classifiers estimate conditional scores with a cosine similarity of $0.9175$ even without self-calibration. This shows that with a well-trained unconditional SGM---in this case, where we use the ground-truth unconditional score---CGSGM is able to produce conditional scores pointing in the correct directions in most cases. This explains why the original CGSGM generates samples with decent quality. After applying self-calibration loss and the scaling method, we further improve the cosine similarity to $0.9689$, which we believe enhances the quality of class-conditional generation.

\section{Classifier-only generation by interpreting classifiers as SGMs}
\label{appendix:cls_only}
\begin{figure*}[t]
\centering
	\begin{subfigure}{0.35\linewidth}
	\includegraphics[width=\linewidth]{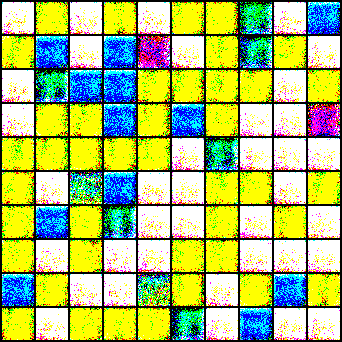}
	\subcaption{Vanilla classifier}
         \label{fig:nocalib_cls_only}
	\end{subfigure}\hspace{0.1\linewidth}
	\begin{subfigure}{0.35\linewidth}
	\includegraphics[width=\linewidth]{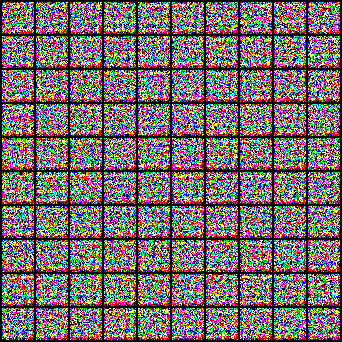}
	\subcaption{DLSM}
         \label{fig:dlsm_cls_only}
	\end{subfigure}
	\begin{subfigure}{0.35\linewidth}
	\includegraphics[width=\linewidth]{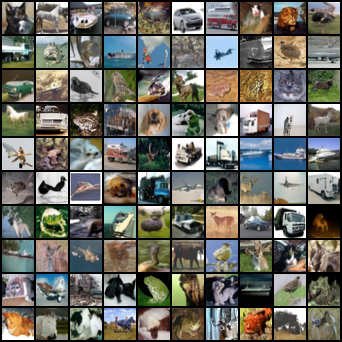}
	\subcaption{Unconditional generation with Self-calibration}
         \label{fig:calib_cls_only_uncond}
	\end{subfigure}\hspace{0.1\linewidth}
	\begin{subfigure}{0.35\linewidth}
	\includegraphics[width=\linewidth]{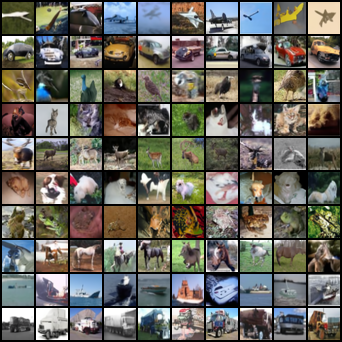}
	\subcaption{Conditional generation with Self-calibration}
         \label{fig:calib_cls_only}
	\end{subfigure}
 
	\caption{Generated images from classifier-only score estimation}
\label{fig:class_only}
\end{figure*}

In this section, we show the results when taking the score estimated by a classifier as an unconditional SGM. For unconditional generation, the classifier is used to estimate the unconditional score; for conditional generation, both terms in Eq.~\ref{eqn:class_guide} are estimated by classifiers. In other words, the time-dependent unconditional score $\nabla_\vx \log p_t(\vx)$ can be written as
\begin{gather}
    \nabla_\vx \log p_t(\vx)=\nabla_\vx \log\Sigma_y\exp f(\vx,y,t;\pmb{\phi})
    \label{eqn:class_only_uncond}
\end{gather}
where $f(\vx,y,t;\pmb{\phi})$ is the logits of the classifier. By adding the gradient of classifier $\nabla_\vx \log p_t(y|\vx)$ to Eq.~\ref{eqn:class_only_uncond}, we obtain the conditional score estimated by a classifier:
\begin{align*}
    \nabla_\vx \log p_t(\vx|y) &= \nabla_\vx \log p_t(\vx) + \nabla_\vx \log p_t(y|\vx) \\
    &= \nabla_\vx \log\Sigma_y\exp f(\vx,y,t;\pmb{\phi}) + \nabla_\vx\log \frac{\exp\left(f(\vx,y,t;\pmb{\phi})\right)}{\sum_y\exp\left(f(\vx,y,t;\pmb{\phi})\right)} \\
    &= \nabla_\vx \log\Sigma_y\exp f(\vx,y,t;\pmb{\phi}) + \nabla_\vx f(\vx,y,t;\pmb{\phi}) - \nabla_\vx \log\Sigma_y\exp f(\vx,y,t;\pmb{\phi}) \\
    &= \nabla_\vx f(\vx,y,t;\pmb{\phi})
\end{align*}
Here, the conditional score is essentially the gradient of the logits. Therefore, we sample from $\nabla_\vx \mathrm{LogSumExp}_y f(\vx,y,t;\pmb{\phi})$ for unconditional generation and $\nabla_\vx f(\vx,y,t;\pmb{\phi})$ for conditional generation.

\begin{table}[h]
\begin{minipage}[c]{1.0\linewidth}
\small\centering
\caption{Quantitative measurements of classifier-only generation with classifier trained using self-calibration loss as regularization}
\label{table:cls_only}
\begin{tabular}{ccccc}
\toprule
Method & FID ($\downarrow$) & IS ($\uparrow$) & intra-FID ($\downarrow$) & Acc ($\uparrow$)\\
\midrule
$\nabla_\vx \log\Sigma_y\exp f(\vx,y,t;\pmb{\phi})$ & 7.54 & 8.93 \\
$\nabla_\vx f(\vx,y,t;\pmb{\phi})$ & 7.26 & 8.93 & 18.86 & 0.890 \\
\bottomrule
\end{tabular}
\end{minipage}
\end{table}

Without self-calibration, both the vanilla classifier (Fig.~\ref{fig:nocalib_cls_only}) and DLSM (Fig.~\ref{fig:dlsm_cls_only}) are unable to generate meaningful images when interpreted as conditional SGMs; this also occurs for unconditional generation. This shows that without related regularization, the interpretation of classifiers as SGMs is not naturally learned through the classification task. After adopting self-calibration loss as regularization, Figures~\ref{fig:calib_cls_only_uncond} and~\ref{fig:calib_cls_only} show that not only does $\nabla_\vx \mathrm{LogSumExp}_y f(\vx,y,t;\pmb{\phi})$ become a more accurate estimator of unconditional score through direct training, $\nabla_\vx f(\vx,y,t;\pmb{\phi})$ also becomes a better estimator of the conditional score as a side effect. Here, we also include the quantitative measurements of unconditional and conditional classifier-only generation in Table~\ref{table:cls_only}.

\section{Tuning the Scaling Factor for Classifier Guidance}
\label{appendix:tune_scale}
This section includes the results when tuning the scaling factor $\lambda_{\mathrm{CG}}$ for classifier guidance with and without self-calibration under the fully-supervised setting.
\begin{figure*}[h]
\begin{minipage}[c]{1.0\linewidth}
	\hspace*{328pt}\begin{subfigure}{0.15\linewidth}
	\includegraphics[width=\linewidth]{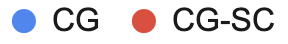}
         \end{subfigure}
\end{minipage}
\centering
	\begin{subfigure}{0.45\linewidth}
	\includegraphics[width=\linewidth]{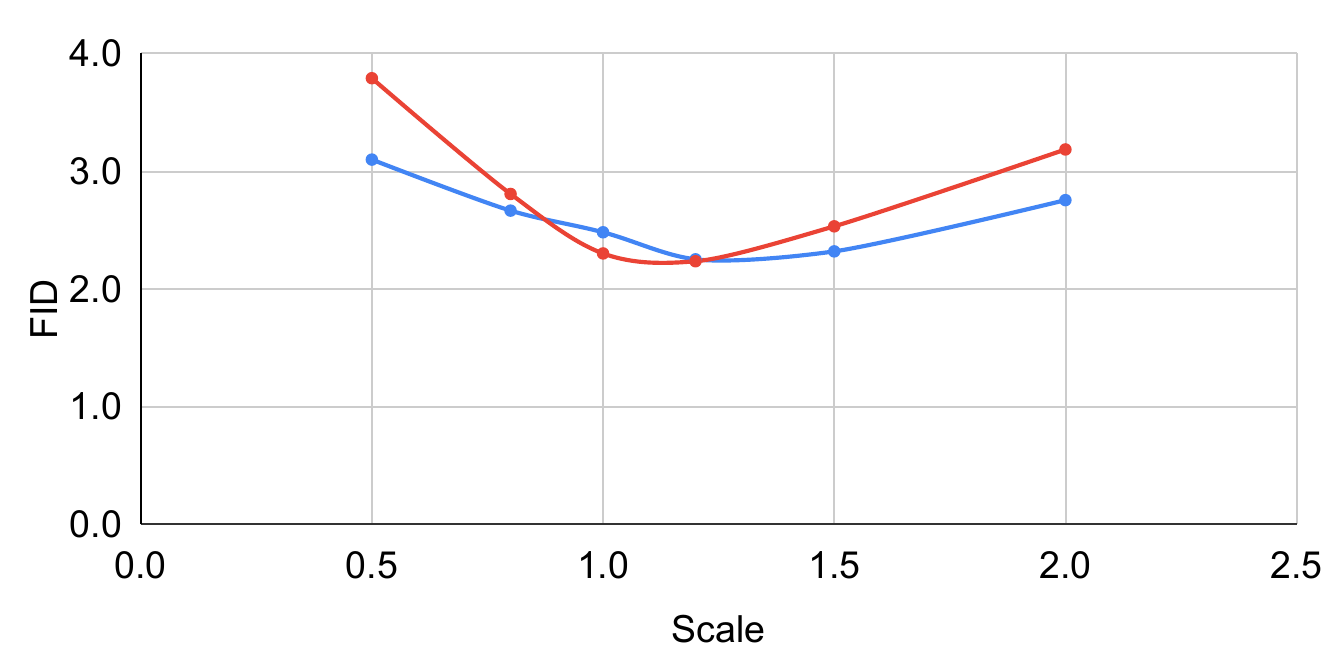}
	\subcaption{FID ($\downarrow$)}
	\end{subfigure}\hspace{0.05\linewidth}
	\begin{subfigure}{0.45\linewidth}
	\includegraphics[width=\linewidth]{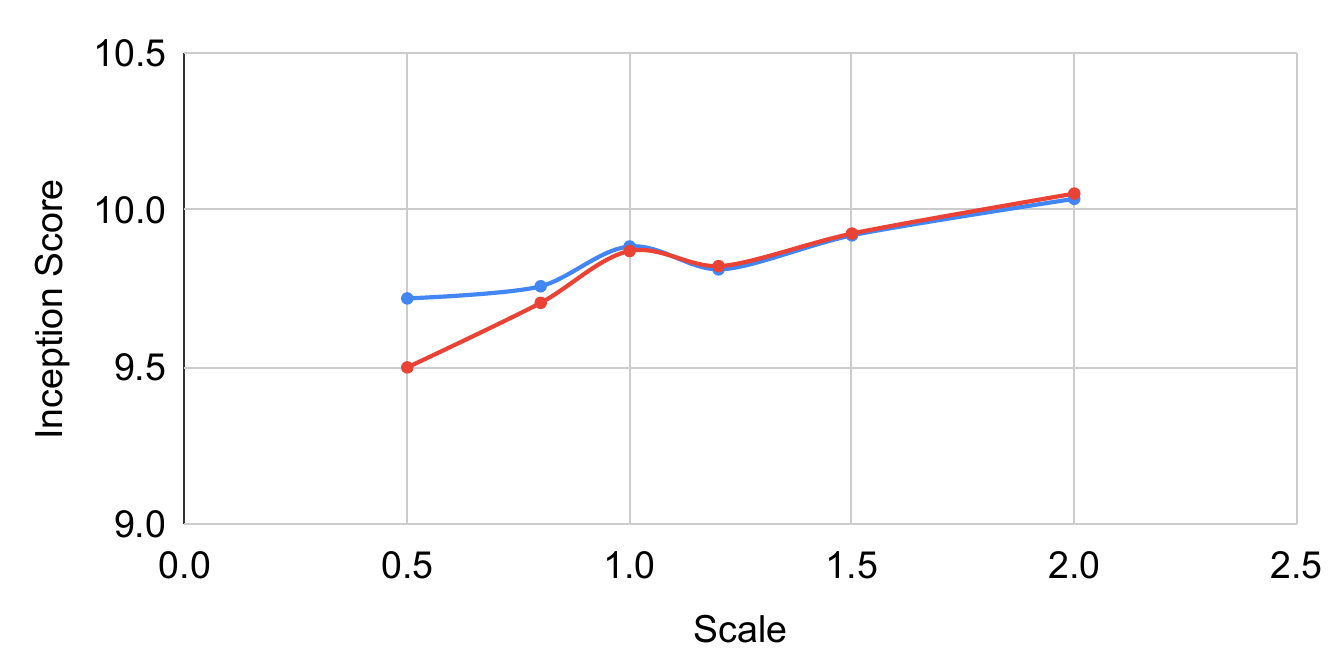}
	\subcaption{Inception Score ($\uparrow$)}
	\end{subfigure}
	\begin{subfigure}{0.45\linewidth}
	\includegraphics[width=\linewidth]{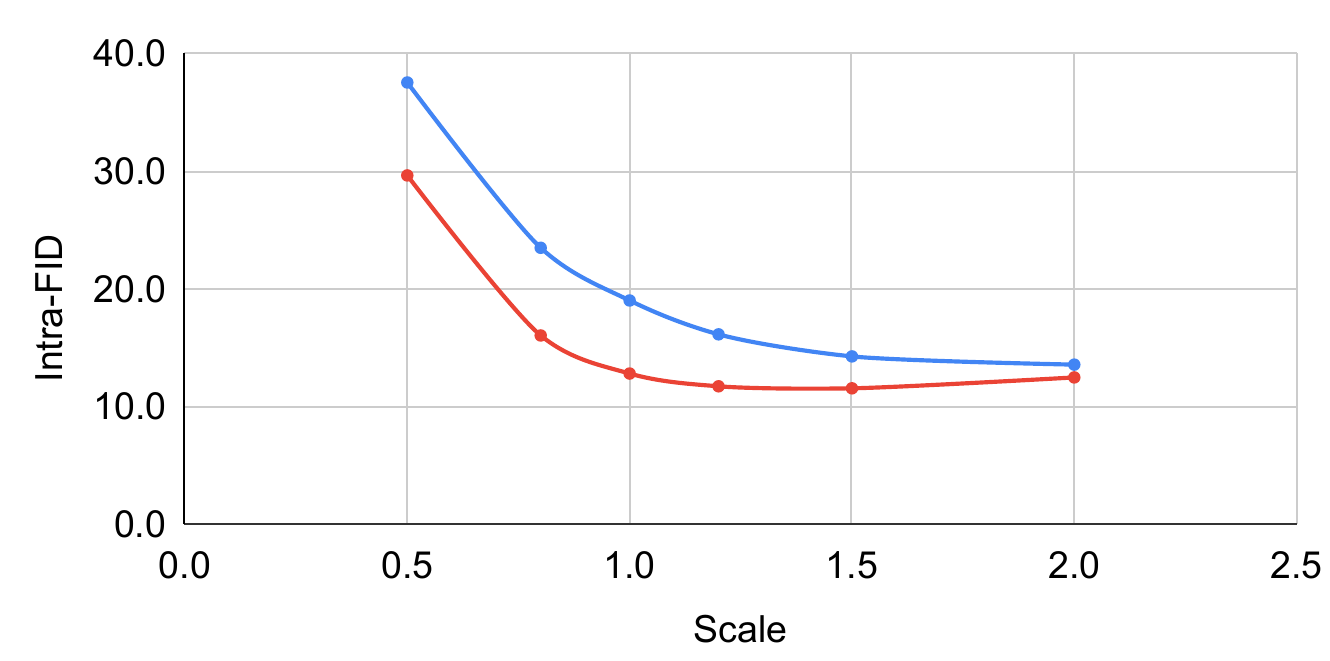}
	\subcaption{intra-FID ($\downarrow$)}
	\end{subfigure}\hspace{0.05\linewidth}
	\begin{subfigure}{0.45\linewidth}
	\includegraphics[width=\linewidth]{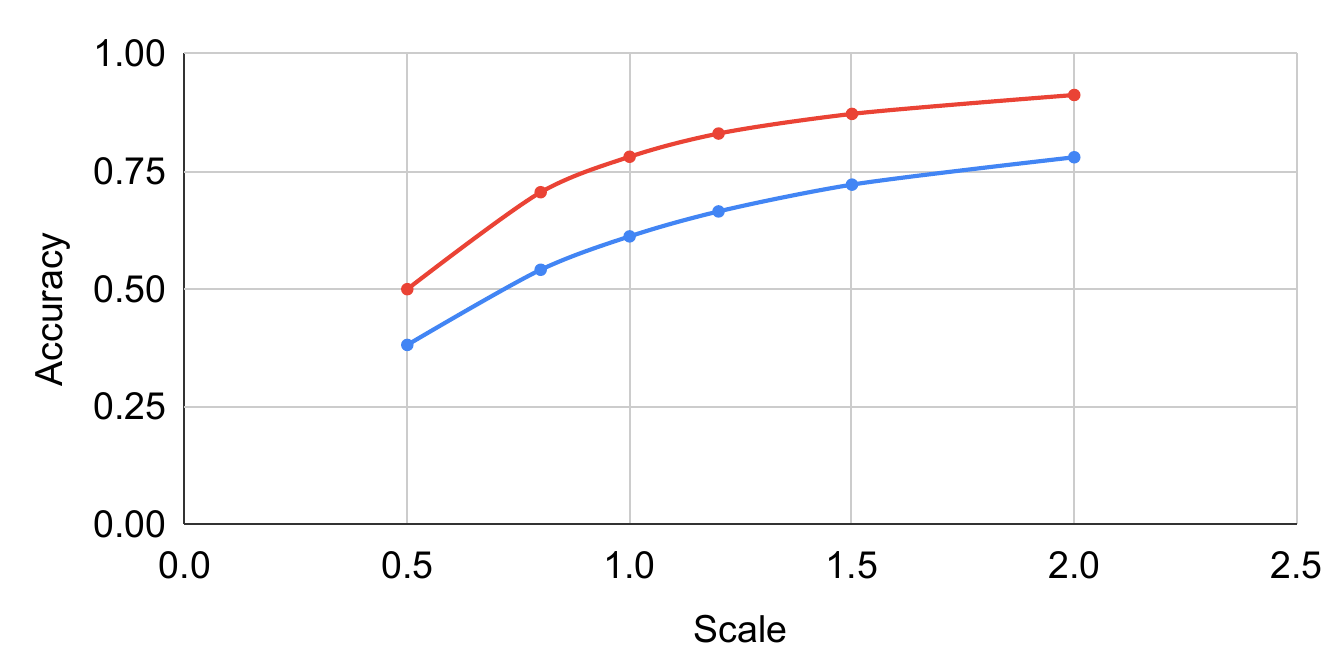}
	\subcaption{Accuracy ($\uparrow$)}
	\end{subfigure}
	\caption{Results when tuning scaling factor $\lambda_{\mathrm{CG}}$ for \textbf{CGSGM} (blue, without self-calibration) and \textbf{CGSGM-SC} (red, with self-calibration). (a)~FID vs.\ $\lambda_{\mathrm{CG}}$. (b)~Inception score vs.\ $\lambda_{\mathrm{CG}}$. (c)~Intra-FID vs.\ $\lambda_{\mathrm{CG}}$. (d)~Generation accuracy vs.\ $\lambda_{\mathrm{CG}}$. Unconditional metrics (FID and IS) differ little, but we observe a distinct performance gap when evaluated conditionally (intra-FID and accuracy).}
\label{fig:fully_tune}
\end{figure*}

Figure~\ref{fig:fully_tune} shows the result when tuning the scaling factor $\lambda_{\mathrm{CG}}$ for classifier guidance. When tuning $\lambda_{\mathrm{CG}}$ with and without self-calibration, self-calibration has little affect on unconditional performance. However, when evaluated with conditional metrics, the improvement after incorporating self-calibration becomes more significant. The improvement in intra-FID reaches $7.9$ whereas generation accuracy improves by as much as $13\%$.

\newpage
\section{Images generated by classifier guidance with and without self-calibration}
\label{appendix:sample}
This section includes images generated by classifier guidance with (first 6~images) and without (last 6~images) self-calibration after training on various percentages of labeled data. Each row corresponds to a class in the CIFAR-10 dataset. Generated images of all method can be found in the supplementary material.

\begin{figure*}[h]
\centering
    \includegraphics[width=0.797\linewidth, frame]{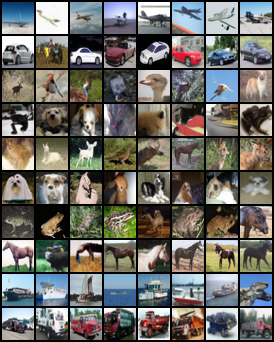}
    \caption{Randomly selected images of classifier guidance with self-calibration ($5\%$ labeled data)}
\end{figure*}

\begin{figure*}[h]
\centering
    \includegraphics[width=0.797\linewidth, frame]{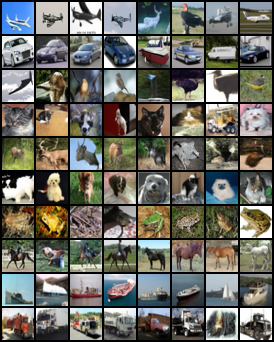}
    \caption{Randomly selected images of classifier guidance with self-calibration ($20\%$ labeled data)}
\end{figure*}

\begin{figure*}[h]
\centering
    \includegraphics[width=0.797\linewidth, frame]{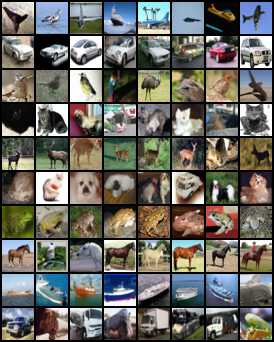}
    \caption{Randomly selected images of classifier guidance with self-calibration ($40\%$ labeled data)}
\end{figure*}

\begin{figure*}[h]
\centering
    \includegraphics[width=0.797\linewidth, frame]{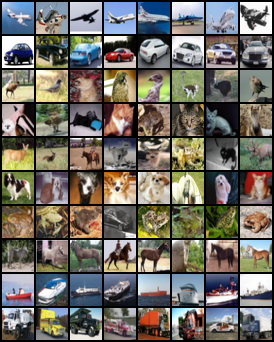}
    \caption{Randomly selected images of classifier guidance with self-calibration ($60\%$ labeled data)}
\end{figure*}

\begin{figure*}[h]
\centering
    \includegraphics[width=0.8\linewidth, frame]{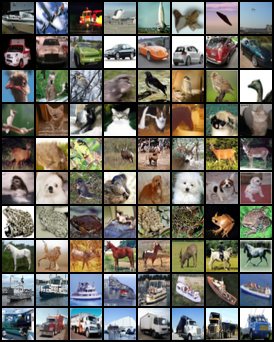}
    \caption{Randomly selected images of classifier guidance with self-calibration ($80\%$ labeled data)}
\end{figure*}

\begin{figure*}[h]
\centering
    \includegraphics[width=0.8\linewidth, frame]{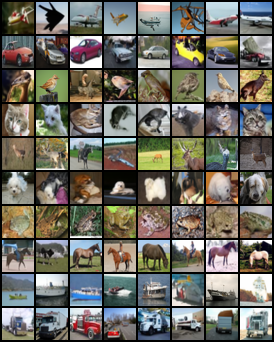}
    \caption{Randomly selected images of classifier guidance with self-calibration ($100\%$ labeled data)}
\end{figure*}

\begin{figure*}[h]
\centering
    \includegraphics[width=0.797\linewidth, frame]{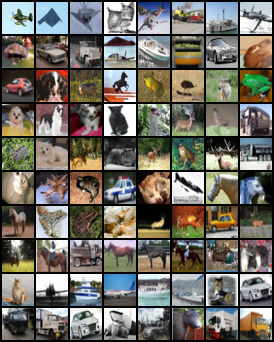}
    \caption{Randomly selected images of vanilla classifier guidance ($5\%$ labeled data)}
\end{figure*}

\begin{figure*}[h]
\centering
    \includegraphics[width=0.797\linewidth, frame]{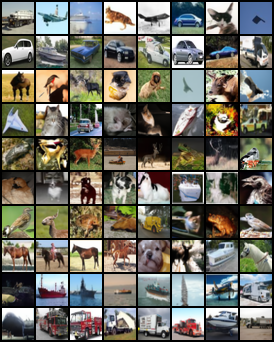}
    \caption{Randomly selected images of vanilla classifier guidance ($20\%$ labeled data)}
\end{figure*}

\begin{figure*}[h]
\centering
    \includegraphics[width=0.797\linewidth, frame]{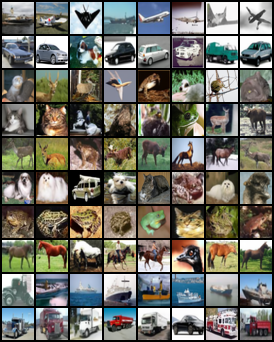}
    \caption{Randomly selected images of vanilla classifier guidance ($40\%$ labeled data)}
\end{figure*}

\begin{figure*}[h]
\centering
    \includegraphics[width=0.797\linewidth, frame]{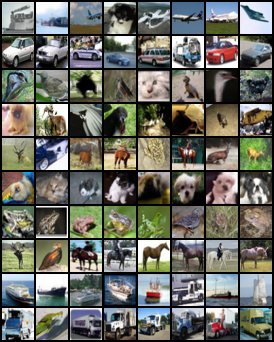}
    \caption{Randomly selected images of vanilla classifier guidance ($60\%$ labeled data)}
\end{figure*}

\begin{figure*}[h]
\centering
    \includegraphics[width=0.8\linewidth, frame]{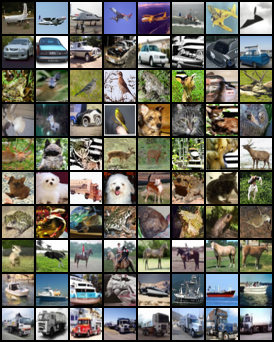}
    \caption{Randomly selected images of vanilla classifier guidance ($80\%$ labeled data)}
\end{figure*}

\begin{figure*}[h]
\centering
    \includegraphics[width=0.8\linewidth, frame]{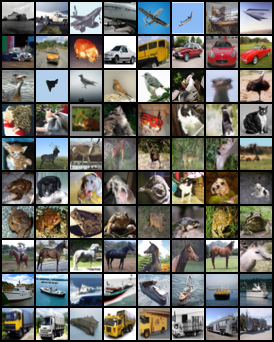}
    \caption{Randomly selected images of vanilla classifier guidance ($100\%$ labeled data)}
\end{figure*}
\end{document}